\pdfoutput=1

\documentclass[10pt,twocolumn,letterpaper]{article}

\usepackage{iccv2023AuthorKit/iccv}
\usepackage[accsupp]{axessibility}
\usepackage{times}
\usepackage{epsfig}
\usepackage{graphicx}
\usepackage{amsmath}
\usepackage{amssymb}
\usepackage{dsfont}
\usepackage{multirow, makecell}
\usepackage{booktabs}
\usepackage{graphbox}
\usepackage{caption}
\usepackage{subcaption}

\DeclareMathOperator*{\argmax}{\mathit{argmax}} 

\usepackage[breaklinks=true,bookmarks=false]{hyperref}

\iccvfinalcopy 

\def\iccvPaperID{12097} 

\begin{document}

\title{
NeSS-ST: Detecting Good and Stable Keypoints with a Neural Stability Score and the Shi-Tomasi detector
}

\author{
Konstantin Pakulev \\
Skolkovo Institute of Science and Technology \\
{\tt\small Konstantin.Pakulev@skoltech.ru} \\
\and
Alexander Vakhitov\\
SLAMcore\\
{\tt\small alexander.vakhitov@gmail.com}
\and
Gonzalo Ferrer\\
Skolkovo Institute of Science and Technology\\
{\tt\small G.Ferrer@skoltech.ru}
}

\maketitle

\begin{abstract}
Learning a feature point detector presents a challenge both due to the ambiguity of the definition of a keypoint and, correspondingly, the need for specially prepared ground truth labels for such points. In our work, we address both of these issues by utilizing a combination of a hand-crafted Shi-Tomasi detector, a specially designed metric that assesses the quality of keypoints, the stability score (SS), and a neural network. We build on the principled and localized keypoints provided by the Shi-Tomasi detector and learn the neural network to select good feature points via the stability score. The neural network incorporates the knowledge from the training targets in the form of the neural stability score (NeSS). Therefore, our method is named NeSS-ST since it combines the Shi-Tomasi detector and the properties of the neural stability score. It only requires sets of images for training without dataset pre-labeling or the need for reconstructed correspondence labels. We evaluate NeSS-ST on HPatches, ScanNet, MegaDepth and IMC-PT demonstrating state-of-the-art performance and good generalization on downstream tasks. The project repository is available at: \url{https://github.com/KonstantinPakulev/NeSS-ST}.


\end{abstract}

\section{Introduction}

Feature point detection (keypoint detection) is usually the first step in camera localization and scene reconstruction pipelines based on sparse features commonly used in robotics~\cite{mur2015orb}, computer vision~\cite{schonberger2016structure}, augmented, mixed and virtual reality~\cite{castle2008video, middelberg2014scalable} and other systems. 

Whilst feature description is successfully approached in the literature as metric learning~\cite{mishchuk2017working, tian2019sosnet} problem, applying deep learning to the feature detection task still poses a challenge with classical solutions showing competitive results on the state-of-the-art benchmarks~\cite{jin2021image}. The difficulty is caused by the innate vagueness of point of interest definition~\cite{witkin1983scalespace} that greatly complicates the formulation of feature detection as a learning problem.

The data required to learn keypoints with sufficient robustness to illumination and viewpoint changes presents another challenge. Structure-from-Motion (SfM)~\cite{schonberger2016structure, dai2017bundlefusion} and Multi-View Stereo (MVS)~\cite{schonberger2016pixelwise, dai2017bundlefusion} reconstructions that are used by some methods~\cite{yi2016lift, ono2018lf, revaud2019r2d2, dusmanu2019d2, tyszkiewicz2020disk} to obtain pixel-accurate correspondences are hard to build properly and lack full image coverage~\cite{li2018megadepth, dai2017scannet}.

We approach the problem of ``defining a keypoint'' by leveraging principled assumptions that make up reasonable keypoints~\cite{shi1994good}: we employ the Shi-Tomasi~\cite{shi1994good} detector to provide locations for keypoints as well as to perform sub-pixel localization.
At the same time, we propose a quantitative metric to measure the stability of detected keypoints to viewpoint transformations, the {\em stability score (SS)}, that randomly perturbs detected points and their surrounding patches for later aggregation of their statistics. This metric can be calculated in an online fashion for a set of keypoints from a single image, so it is ideal for automatically generating a supervised signal for training a neural network that predicts the {\it neural stability score (NeSS)}. We empower a classical method, the {\bf S}hi-{\bf T}omasi detector, with a neural approach, NeSS, getting a method, NeSS-ST, that provides accurate locations of keypoints that are more likely to remain stable under viewpoint changes.

The design presented in our work does not rely on ground-truth poses or reconstructed correspondences alongside~\cite{detone2018superpoint, barroso2019key, lee2022self}. These methods are sometimes referred to as self-supervised~\cite{detone2018superpoint, lee2022self}. Compared to the prior art, NeSS-ST needs only a set of real images without any labels for training. Out of self-supervised methods, only Key.Net~\cite{barroso2019key} and REKD~\cite{lee2022self} work in a similar setting while SuperPoint~\cite{detone2018superpoint} relies on pre-labeled datasets and requires pre-training a base detector on a synthetic dataset (see Table~\ref{tab:self_supervised}).

Our contributions are as follows. We present a novel metric, the stability score, to estimate the quality of keypoints. We propose a method, NeSS-ST, that employs the stability score to learn a neural network and requires only images for training. In terms of pose accuracy, NeSS-ST surpasses state-of-the-art on MegaDepth~\cite{li2018megadepth}, is on par on IMC-PT~\cite{jin2021image}, ScanNet~\cite{dai2017scannet} and HPatches~\cite{balntas2017hpatches} being the best among the self-supervised methods (see Table~\ref{tab:overall_ranking}).

\begin{table}[thpb]
\begin{center}
\resizebox{\columnwidth}{!}{
\begin{tabular}{lccc}
\toprule
\multirow{2}{*}{Methods} & \multirowcell{2}{Pre-training \\ base detector} & \multirowcell{2}{Dataset \\ pre-labeling} & \multirowcell{2}{Ground-truth \\ generation} \\
\\
\midrule
SuperPoint~\cite{detone2018superpoint} & Yes & Yes & Offline \\
Key.Net~\cite{barroso2019key} & No & No & Offline  \\
REKD~\cite{lee2022self} & No & No & Offline  \\
\midrule
NeSS-ST & No & No & Online  \\
\bottomrule
\end{tabular}
}
\end{center}
\caption{Comparison of self-supervised methods for learning keypoints.}
\label{tab:self_supervised}
\end{table}

\section{Related work}
\label{sec:related_work}

{\bf Handcrafted detectors}. 
Since Moravec~\cite{moravec1981rover}, feature detection methods focused on finding local extrema in signals derived from images that correspond to meaningful structures, \eg corners~\cite{witkin1983scalespace}. Therefore, the earliest designs of detectors employ the grayvalue analysis via differential expressions~\cite{witkin1983scalespace, harris1988combined, blostein1989multiscale, shi1994good, lindeberg1998feature}.

Depending on image resolution or scale, both location and strength of keypoints response change \cite{witkin1983scalespace}, thus, spatial extrema do not necessarily constitute good features. The most productive way to tackle this problem is modeling scale via the scale-space~\cite{lindeberg1998feature} - the backbone of some famous methods like SIFT~\cite{lowe2004distinctive}, SURF~\cite{bay2006surf}, Harris-Laplace~\cite{mikolajczyk2001indexing}, Harris-Affine~\cite{mikolajczyk2002affine}, KAZE~\cite{alcantarilla2012kaze}. Features are detected either as extrema in the scale-space~\cite{lowe2004distinctive, bay2006surf, alcantarilla2012kaze}, or extrema are found at scales first, and then features are selected among extrema by using transformation invariant scores~\cite{mikolajczyk2001indexing, mikolajczyk2002affine}.

{\bf Learned detectors}. Existing designs of learned detectors are quite diverse. The majority of the methods employ correspondence labels for training which are obtained via SfM and MVS~\cite{yi2016lift, ono2018lf, dusmanu2019d2, tyszkiewicz2020disk} or optical flow~\cite{revaud2019r2d2}. Preparing dense pixel-accurate ground-truth correspondence labels poses the problem by itself~\cite{schonberger2016pixelwise, dai2017bundlefusion} - special pre-processing of data and validation of obtained results is required~\cite{revaud2019r2d2, dai2017scannet, li2018megadepth}. Although avoiding surface reconstruction can ease the problem formulation~\cite{schonberger2016structure, schonberger2016pixelwise}, accurate poses can still be hard to obtain~\cite{zhang2021survey}. In this regard, self-supervised methods that do not require reconstructed correspondences, such as~\cite{detone2018superpoint, barroso2019key, lee2022self} or the method proposed here, present a viable alternative.

Self-supervised methods rely on sampling of homographies~\cite{detone2018superpoint}, affinities~\cite{barroso2019key} and in-plane rotations~\cite{lee2022self} to generate ground truth correspondences. To detect keypoint candidates SuperPoint~\cite{detone2018superpoint} employs a base detector that is trained on a synthetic dataset with corner-like structures. Key.Net~\cite{barroso2019key} and REKD~\cite{lee2022self} follow an anchorless approach for detection: keypoints are discovered as a result of a loss function optimization. To address the problem of scale SuperPoint~\cite{detone2018superpoint} prepares ground truth feature points using accumulated repeatability-like scores obtained with the help of the base detector and generated correspondences. Key.Net~\cite{barroso2019key} and REKD~\cite{lee2022self}, on the other hand, average values of the loss function over different window sizes. 

In our method, similarly to~\cite{mikolajczyk2001indexing, mikolajczyk2002affine}, we find extrema at a scale using the detector based on the structure tensor~\cite{harris1988combined, shi1994good}, the Shi-Tomasi~\cite{shi1994good} detector (see Sec.~\ref{sec:shi} and Sec.~\ref{sec:base_detector_abl}), and perform feature selection using the scores predicted by the neural network that assess the stability of points to viewpoint changes. Like SuperPoint~\cite{detone2018superpoint}, we utilize a combination of a base detector and homography generation to calculate ground truth scores. We leverage a specially designed keypoint quality measure that, unlike SuperPoint, assesses feature points locally, using only a small neighbourhood (see Sec.~\ref{sec:kss} and Sec.~\ref{sec:stability_score_abl}). That, coupled with the handcrafted detector, allows us to prepare the required ground truth during the training using the same GPU (see Table~\ref{tab:self_supervised}, Sec.~\ref{sec:training} and Sec.~\ref{sec:implementation}).

\section{Method}

\begin{figure*}
\begin{center}
\includegraphics[align=c, width=\linewidth]{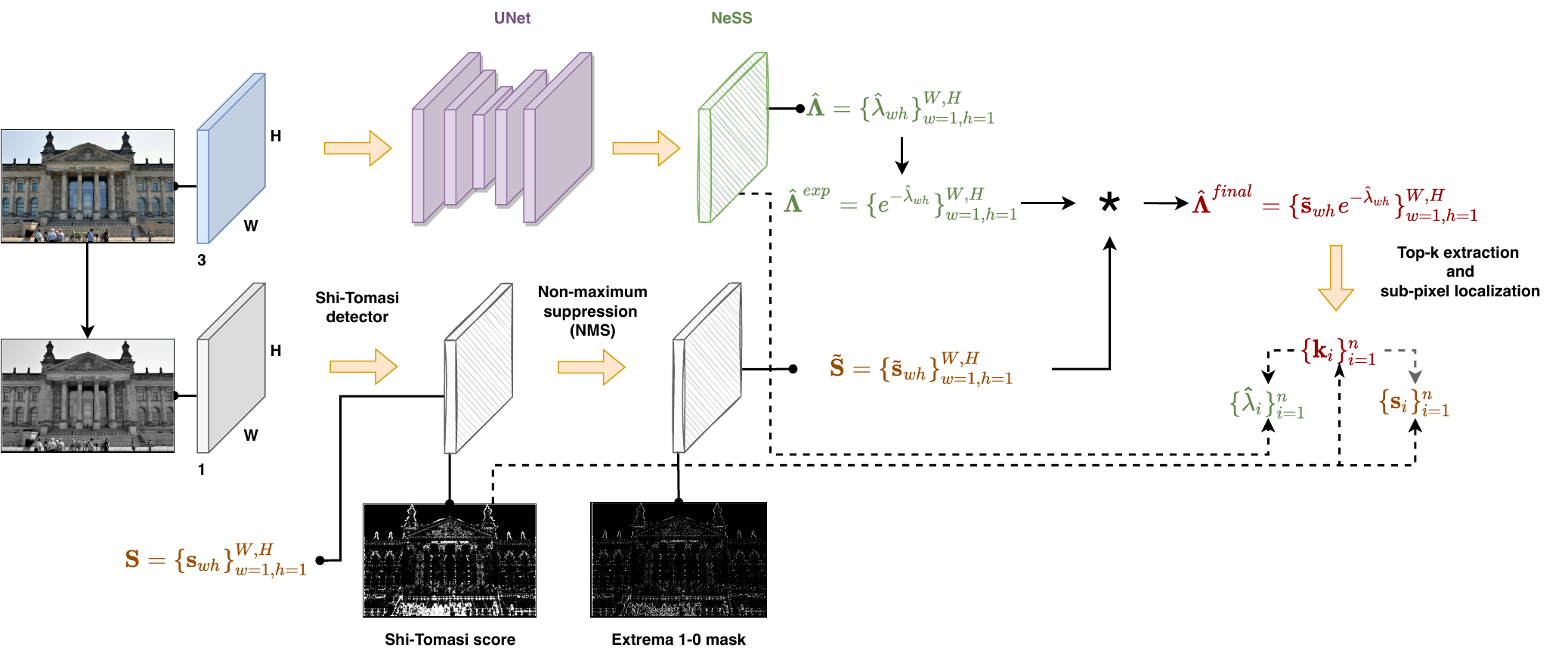}
\end{center}
   \caption{The method applies the Shi-Tomasi detector to an input image to get the Shi-Tomasi score $\mathbf{S}$. Next, a binary mask of extrema $\tilde{\mathbf{S}}$ is obtained via non-maximum suppression of $\mathbf{S}$. Simultaneously, the method uses the neural network to regress the neural stability score $\hat{\mathbf{\Lambda}}$.
   A set of best feature points $\{\mathbf{k}_i\}^n_{i=1}$ is selected from the combined score map $\hat{\mathbf{\Lambda}}^{\mathit{final}}$ that is a multiplicative combination of $\tilde{\mathbf{S}}$ with the negative exponential of $\hat{\mathbf{\Lambda}}$. Obtained keypoints are localized using $\mathbf{S}$ and are provided with corresponding $\{\mathbf{s}_i\}^n_{i=1}$ and $\{\hat{\mathbf{\lambda}}_i\}^n_{i=1}$.}
\label{fig:method}
\end{figure*}

NeSS-ST is the combination of the handcrafted Shi-Tomasi~\cite{harris1988combined, shi1994good} detector and the neural network (see Fig.~\ref{fig:method}). We briefly present the formulation of the Shi-Tomasi detector as well as describe the sub-pixel localization procedure in Sec.~\ref{sec:shi}. In Sec.~\ref{sec:kss} we describe the measure that allows us to quantitatively assess keypoints, the {\em stability score (SS)}. Sec.~\ref{sec:training} discusses how the stability score can be used to prepare the ground truth for training the neural network to predict the stability score for good feature points, the {\em neural stability score (NeSS)}. Finally, Sec.~\ref{sec:implementation} gives an account of implementation details.

\subsection{Shi-Tomasi detector}
\label{sec:shi}

We use the Shi-Tomasi~\cite{harris1988combined, shi1994good} detector to get locations of feature points (see Sec.~\ref{sec:base_detector_abl} for the motivation behind the choice). We calculate the second-moment matrix for each pixel using the Gaussian weighting function~\cite{harris1988combined, shi1994good} and assign the pixel with a score that is equal to the smallest eigenvalue of the second-moment matrix~\cite{shi1994good}. By applying the non-maximum suppression over the obtained score map, we get the locations of keypoints.

We further refine the locations of feature points following~\cite{lowe2004distinctive}. By assuming that the Shi-Tomasi score $\mathbf{S}$ can be approximated by a function $\mathcal{S}(\mathbf{x})$, we perform the second order Taylor expansion around a point $\mathbf{x}$  and find the correction $\mathbf{dx}$ that maximizes $\mathcal{S}(\mathbf{x})$ by solving:

\begin{equation}
\mathbf{dx}=-{\frac{\partial^2 \mathcal{S}}{{\partial \mathbf{x}}^2}}^{-1}\frac{\partial \mathcal{S}}{\partial \mathbf{x}}.
\end{equation}

\subsection{Stability score}
\label{sec:kss}

In this work, we design a quantitative description to assess the stability of feature points to viewpoint transformations. Given a keypoint $\mathbf{k}$ we apply to it a set of generated homographies $\{\mathcal{H}_j\}^{m}_{j=1}$ to produce a set of $m$ warped keypoints $\{\mathbf{k}^{\prime}_j\}^m_{j=1}$:

\begin{equation}
\mathbf{k}^{\prime}_j = \mathcal{H}_j\mathbf{k}.
\end{equation}

We generate homographies by sampling random perspective distortions while restricting the deformation along the x and y axes, see the details in the supplementary material. We do not add rotation or translation transformations since the Shi-Tomasi
detector with Gaussian weighting that we use is invariant to them~\cite{harris1988combined, shi1994good, dufournaud2000matching}.

Next, we create a grid $\mathbf{G}_j$ of the predetermined size $p$ around each of the warped points $\mathbf{k}^{\prime}_j$, warp the grid back to the image coordinates and sample values in these locations to get a deformed patch $\mathbf{P}_j$ around $\mathbf{k}^{\prime}_j$:

\begin{equation}
\mathbf{P}_j = \mathit{sample}(\mathcal{H}^{-1}_j\mathbf{G}_j).
\end{equation}

The Shi-Tomasi detector is not invariant to scale~\cite{dufournaud2000matching} or perspective transformations, hence, in general, we cannot accumulate its scores from different warped patches~\cite{lindeberg1998feature}. However, we can accumulate locations of maximum scores and analyze their distribution.

We run the Shi-Tomasi detector $f_\mathit{Shi}$ on each patch $\mathbf{P}_j$ getting a score patch $\mathbf{S}^{\mathit{patch}}_j = \{\mathbf{s}^{\mathit{patch}}_l\}^{p^2}_{l=1}$ and extract from it the location $\hat{\mathbf{l}}_j$ that corresponds to the maximum score:

\begin{equation}
\label{eq:shi_argmax}
\hat{\mathbf{l}}_j = \argmax_{\mathbf{s}^{\mathit{patch}}_l} f_\mathit{Shi}(\mathbf{P}_j).
\end{equation}

Next, we warp each $\hat{\mathbf{l}}_j$ back to the original reference frame and calculate the sample covariance $\mathbf{\Sigma}$ with respect to the original position $\mathbf{k}$:

\begin{equation}
\label{eq:sigma}
\mathbf{\Sigma} = \sum^m_{j=1} \frac{(\mathcal{H}^{-1}_j\hat{\mathbf{l}}_j - \mathbf{k})(\mathcal{H}^{-1}_j\hat{\mathbf{l}}_j - \mathbf{k})^{\top}}{m}.
\end{equation}

Finally, we characterize a feature point by the largest of eigenvalues $\mathbf{\lambda}_1$ and $\mathbf{\lambda}_2$ of $\mathbf{\Sigma}$:

\begin{equation}
\label{eq:lambda}
\mathbf{\lambda} = \max(\mathbf{\lambda}_1, \mathbf{\lambda}_2) = \|\Sigma\|_2.
\end{equation}

The stability score $\mathbf{\lambda}$ captures the maximum deviation of a point from its location under perspective transformations of the neighbourhood of the point. The measure prioritizes feature points that deviate the least from the initial location $\mathbf{k}$ in any direction and, hence, are more likely to be accurately detected and localized. For further discussion of the design of the stability score see Sec.~\ref{sec:stability_score_abl}.

Since $\mathbf{\lambda}$ assesses keypoints based on their local perturbations, the patch size $p$ is set to be small. It allows to quickly calculate $\mathbf{\lambda}$ with a sufficient number of samples $m$ for a reasonable number of keypoints $n$. This particular trait makes a foundation for the training process of the NeSS regression network which is discussed next.

\subsection{Neural stability score}
\label{sec:training}

Not every point with a good stability score $\mathbf{\lambda}$ makes a good training target: given that $\mathbf{\lambda}$ is calculated over a small window, image artifacts in low-textured regions can be perfectly stable. These unreliable responses can be filtered by the Shi-Tomasi detector assuming a threshold $t_\mathit{Shi}$. Thus, the neural network is trained on feature points that both have good stability scores and are more likely to correspond to meaningful patterns in an image. This gives rise to a new, better, kind of score, the neural stability score, see Sec.~\ref{sec:stability_score_abl} for more details. In this regard, the stability score can be viewed as the necessary (but not sufficient) condition for a good keypoint. See the supplementary material for results on the influence of $t_\mathit{Shi}$ on the quality of the model.

In line with several works~\cite{ono2018lf, barroso2019key, tyszkiewicz2020disk}, we get keypoints for training the neural network by running keypoint extraction using the trained-so-far weights of the model. More specifically, for each image we extract $n$ feature points $\{\mathbf{k}_i\}^n_{i=1}$, their Shi-Tomasi scores $\{\mathbf{s}_i\}^n_{i=1}$ and neural stability scores $\{\hat{\mathbf{\lambda}}_i\}^n_{i=1}$ regressed by the neural network (see Fig. \ref{fig:method}). Such an approach not only narrows the gap between train and test regimes of the neural network, but it also allows to significantly reduce the time for preparing the ground truth. The combination of during-the-training feature extraction and fast calculation of $\mathbf{\lambda}$ 
enables online ground truth generation.

For each point $\mathbf{k}_i$ we calculate the ground-truth stability score $\mathbf{\lambda}_i$ (Eq.~\ref{eq:lambda}). Next, we define $\mathds{1}(\mathbf{s}_i, t_\mathit{Shi})$ as an indicator function that gives $1$ if $\mathbf{s}_i > t_\mathit{Shi}$ and $0$ otherwise. Finally, we learn $\{\hat{\mathbf{\lambda}}_i\}^n_{i=1}$ by formulating the training objective as a regression problem:

\begin{equation}
\label{eq:ss_loss}
L = 0.5\frac{\sum^n_{i=1} (\hat{\mathbf{\lambda}}_i -  \mathbf{\lambda}_i)^2 \mathds{1}(\mathbf{s}_i, t_\mathit{Shi})}{\sum^n_{i=1}\mathds{1}(\mathbf{s}_i, t_\mathit{Shi})}.
\end{equation}

\subsection{Implementation details}
\label{sec:implementation}

We train our detector from scratch on the same subset of the MegaDepth~\cite{li2018megadepth} dataset as used in DISK~\cite{tyszkiewicz2020disk}. We do not use poses, depth maps or any other information from the dataset other than images. We use full-resolution images cropped to a square of length $560$ pixels for training. We perform validation and model selection on the validation subset of the IMC-PT~\cite{jin2021image} dataset.

We set $p=5$ to correspond to the size of the non-maximum suppression kernel. The value of $n=1024$ was found to be enough to cover the most of salient points above $t_{\mathit{Shi}}$ for our choice of the threshold and resolution of training images. We pick $m=100$ as it provides consistent estimates of $\mathbf{\lambda}$. Further increasing $m$ gives diminishing returns.

We use a U-Net~\cite{ronneberger2015u} architecture with 4 down-sampling layers and $3 \times 3$ convolutions. To train the model we employ Adam~\cite{kingma2015adam} optimizer with learning rate $10^{-4}$. The model that is used in experiments in Sec.~\ref{sec:experiments} was trained on a single NVIDIA 2080 Ti GPU for 22 hours.

\section{Experiments}
\label{sec:experiments}

The means of detector evaluation deserve special attention. Originally, detectors were mostly evaluated using classical metrics like repeatability and matching score~\cite{mikolajczyk2005comparison}. Since feature points require supplying them with a description in order to obtain the correspondences, a descriptor has to be accounted for to ensure a proper assessment of the quality of detected keypoints. A common solution is to fix a descriptor for all detectors in the evaluation~\cite{mikolajczyk2005comparison, verdie2015tilde, zhang2017learning, barroso2019key}. More recent publications reported evidence that gains in classical metrics do not necessarily translate to the gains in the downstream tasks performance~\cite{jin2021image}. For this reason, in our work, we mostly perform the evaluation on a range of downstream tasks. To get a more comprehensive assessment, we test our method on a variety of datasets and tasks.

In our evaluation, we consider the following methods: Shi-Tomasi~\cite{harris1988combined, shi1994good}, SIFT~\cite{lowe2004distinctive}, SuperPoint~\cite{detone2018superpoint}, R2D2~\cite{revaud2019r2d2}, Key.Net~\cite{barroso2019key}, DISK~\cite{tyszkiewicz2020disk} and REKD~\cite{lee2022self}. Some of these methods use detectors that have assigned names, \eg SIFT utilizes the difference of Gaussians (DoG) detector, while some don't, \eg the detector of R2D2 doesn't have a name. Since we report the evaluation results for "detector+descriptor" pairs, for consistency, we refer to both detectors and descriptors by the name of a complete solution that they belong to. For example, "SIFT+DISK" implies that the detector of SIFT, DoG, is used with the descriptor of DISK.

Image resolution plays an important role in the accuracy of correspondences, thus we provide images in the original resolution to all methods in our evaluation. It also allows to assess the ability of a detector to generalize beyond the training resolution. The number of keypoints that is extracted from each image plays no lesser role: we found the regime of 2048 keypoints per image from~\cite{jin2021image, tyszkiewicz2020disk} to be a good trade-off between performance and consumption of computational resources. We use mutual nearest neighbour matching and employ the Lowe ratio test~\cite{lowe2004distinctive, jin2021image}. As choosing proper hyper-parameters for a method is of utmost importance for downstream tasks~\cite{jin2021image}, we employ a hyper-parameter tuning procedure similar to one in~\cite{jin2021image}. We extract feature points for all methods using only the original scale with the exception of Key.Net~\cite{barroso2019key} and REKD~\cite{lee2022self} that have multi-scaling as an essential built-in part of the method. To ensure a fair comparison, we do not employ the orientation estimation for REKD~\cite{lee2022self} and SIFT~\cite{lowe2004distinctive} since other methods in the evaluation don't have it. As long as the sub-pixel localization is a part of our solution, the version of the Shi-Tomasi~\cite{harris1988combined, shi1994good} detector employed in our evaluation includes it since we focus on assessing the influence of NeSS. 

\subsection{Evaluation on HPatches}

The HPatches~\cite{balntas2017hpatches} dataset features sequences with planar surfaces related by homographies under a variety of illumination and viewpoint changes. We use the same test subset as in~\cite{dusmanu2019d2} totaling 540 image pairs from 108 scenes among which 260 image pairs are with illumination changes and 280 - with viewpoint.

We report Mean Matching Accuracy (MMA)~\cite{mikolajczyk2005performance, dusmanu2019d2} under different pixel thresholds following~\cite{detone2018superpoint, dusmanu2019d2, wang2020learning,tyszkiewicz2020disk, lee2022self}. We evaluate on a downstream task of homography estimation using a protocol similar to~\cite{detone2018superpoint, wang2020learning} and report the homography estimation accuracy for different pixel thresholds. Additionally, we integrate the accuracy-threshold curve up to a 5-pixel threshold to get a single-valued quality measure, mean Average Accuracy~\cite{yi2018learning, jin2021image}.

The DISK~\cite{tyszkiewicz2020disk} descriptor shows the state-of-the-art performance on HPatches, thus we pick it for this dataset. We employ OpenCV~\cite{bradski2000opencv} routines for homography estimation. For tuning hyper-parameters we use sequences of HPatches~\cite{balntas2017hpatches} left out from the test set~\cite{dusmanu2019d2} as well as hyper-parameters obtained from validation sequences of IMC-PT~\cite{jin2021image} on the task of relative pose estimation. This combination provides the best results for all methods, see the details in the supplementary material.

\begin{table*}
\begin{center}
\begin{tabular}{lccc}
\toprule
\multirow{2}{*}{Methods} & \multirowcell{2}{Overall \\ mAA (5px)} & \multirowcell{2}{Illumination \\ mAA (5px)} & \multirowcell{2}{Viewpoint \\ mAA (5px)} \\
\\
\midrule
Shi-Tomasi~\cite{harris1988combined, shi1994good} + DISK~\cite{tyszkiewicz2020disk} & \textbf{\textcolor[HTML]{B6321C}{0.716}} & \textbf{\textcolor[HTML]{3C8031}{0.892}} & \textbf{\textcolor[HTML]{3C8031}{0.552}} \\
SIFT~\cite{lowe2004distinctive} + DISK~\cite{tyszkiewicz2020disk} & 0.688 & 0.877 & 0.512 \\
SuperPoint~\cite{detone2018superpoint} + DISK~\cite{tyszkiewicz2020disk} & \textbf{\textcolor[HTML]{006795}{0.706}} & 0.883 & 0.541 \\
R2D2~\cite{revaud2019r2d2} + DISK~\cite{tyszkiewicz2020disk} & 0.690 & \textbf{\textcolor[HTML]{006795}{0.888}} & 0.506 \\
Key.Net~\cite{barroso2019key} + DISK~\cite{tyszkiewicz2020disk} & 0.678 & 0.844 & 0.524 \\
DISK~\cite{tyszkiewicz2020disk} & 0.699 & 0.867 & \textbf{\textcolor[HTML]{006795}{0.542}} \\
REKD~\cite{lee2022self} + DISK~\cite{tyszkiewicz2020disk} & 0.689 & \textbf{\textcolor[HTML]{B6321C}{0.895}} & 0.498 \\
\midrule
NeSS-ST + DISK~\cite{tyszkiewicz2020disk} & \textbf{\textcolor[HTML]{3C8031}{0.714}} & 0.883 & \textbf{\textcolor[HTML]{B6321C}{0.556}} \\
\bottomrule
\end{tabular}
\end{center}
\caption{Evaluation on HPatches~\cite{balntas2017hpatches} with 2048 keypoints and full resolution images. We report homography estimation mAA~\cite{detone2018superpoint, wang2020learning, yi2018learning, jin2021image} up to a 5-pixel threshold. Best results are marked in \textbf{\textcolor[HTML]{B6321C}{red}}, 2nd best - in \textbf{\textcolor[HTML]{3C8031}{green}}, 3rd best - in \textbf{\textcolor[HTML]{006795}{blue}}.}
\label{tab:hpatches}
\end{table*}

\begin{figure*}
\begin{center}
\includegraphics[width=0.95\linewidth]{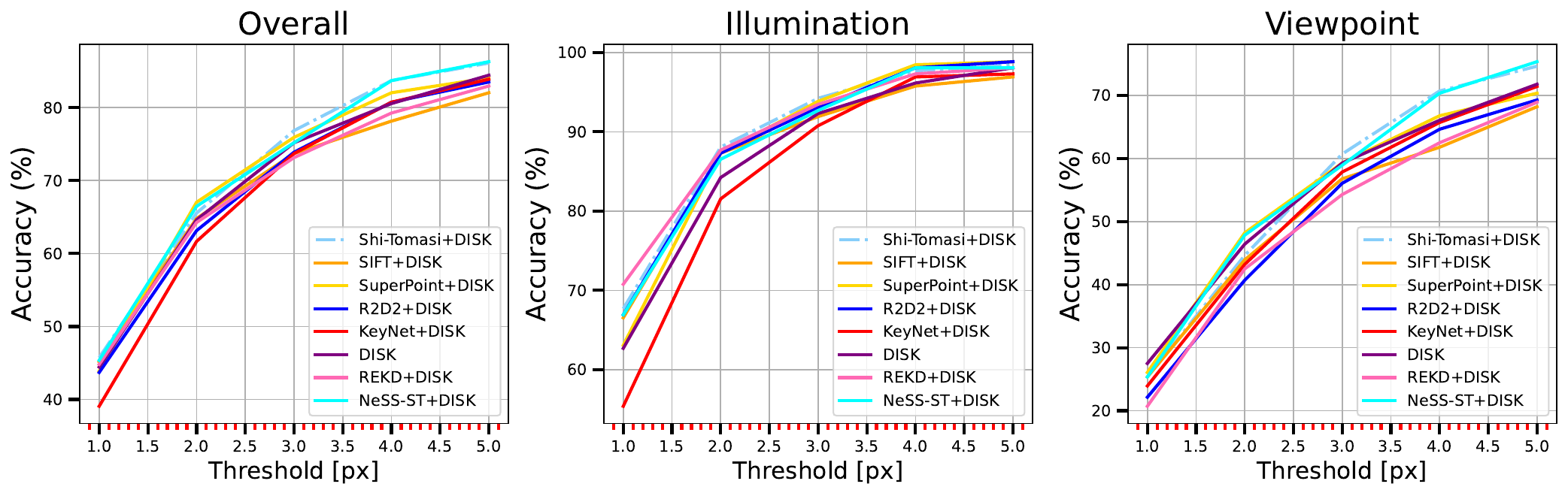}
\end{center}
   \caption{Evaluation on HPatches~\cite{balntas2017hpatches} with 2048 keypoints and full resolution images. We report homography estimation accuracy~\cite{detone2018superpoint, wang2020learning} in \%.}
\label{fig:hpatches_homography}
\end{figure*}

\begin{figure*}
\begin{center}
\includegraphics[width=0.95\linewidth]{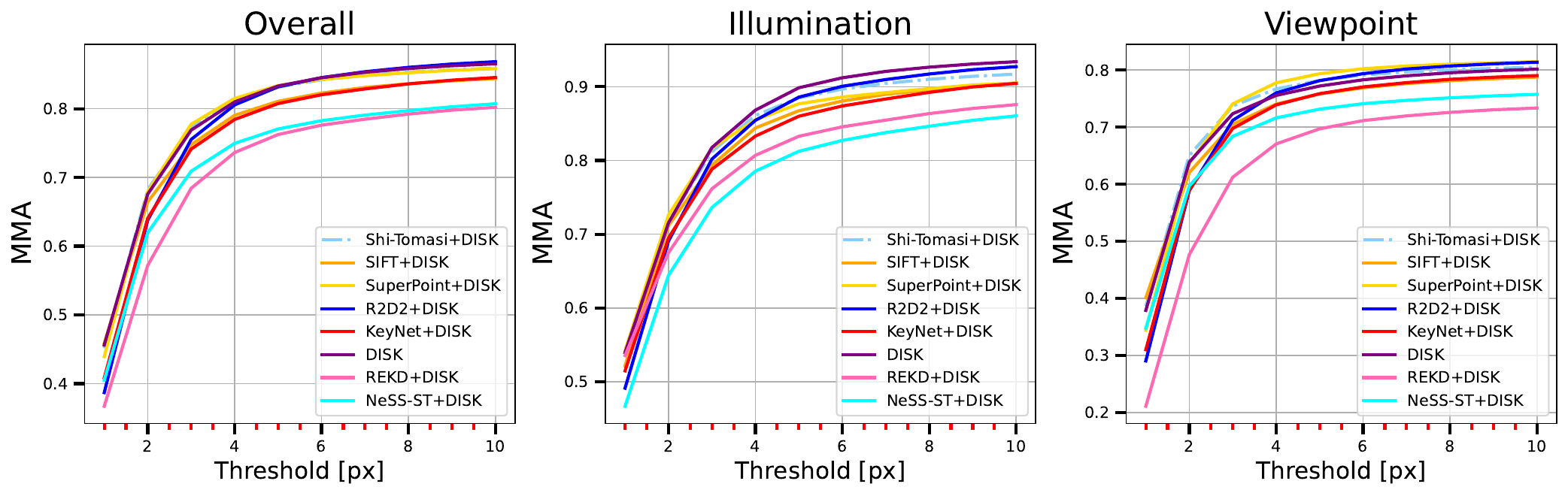}
\end{center}
   \caption{Evaluation on HPatches~\cite{balntas2017hpatches} with 2048 keypoints and full resolution images. We report MMA~\cite{mikolajczyk2005performance, dusmanu2019d2}.}
\label{fig:hpatches}
\end{figure*}

\paragraph{Results.} Our method shows inferior performance to the Shi-Tomasi~\cite{harris1988combined, shi1994good} detector when evaluated on the MMA metric as illustrated in Fig.~\ref{fig:hpatches}. However, the evaluation on the task of homography estimation completely changes the ranking, and our method shows strong performance on scenes with viewpoint changes (see Table~\ref{tab:hpatches} and Fig.~\ref{fig:hpatches_homography}). These results correlate with recent findings that classical methods can show state-of-the-art performance if tuned properly, and that classical metrics might not fully capture the complicated dependency between features and downstream tasks~\cite{jin2021image}. Since we do not address the problem of illumination invariance in our work, NeSS-ST shows average results on illumination sequences. Refer to the supplementary material for more results.

\subsection{Evaluation on downstream tasks}

We evaluate on a downstream task of relative pose estimation following the protocol of~\cite{jin2021image}. We use our own evaluation pipeline to provide consistency in evaluations across different datasets. We calculate pose estimation accuracy for different angular thresholds and report mean Average Accuracy (mAA)~\cite{yi2018learning, jin2021image} for both rotation and translation by integrating the area under the accuracy-threshold curve up to a 10-degree threshold. Errors for rotation and translation are calculated in degrees~\cite{yi2018learning, wang2020learning, jin2021image}.

\subsubsection{Evaluation on IMC-PT}
The IMC-PT~\cite{jin2021image} dataset is a collection of photo-tourism images supplied with depth maps and poses, which are reconstructed via SfM and MVS, that features landmarks (mostly buildings). We use a full test set release that consists of 800 unique images from 8 different locations. By considering image pairs with co-visibility larger than 0.1~\cite{jin2021image}, we get 37k test image pairs.

Like on HPatches, we choose the DISK~\cite{tyszkiewicz2020disk} descriptor as it shows the state-of-the-art performance on this dataset. We utilize a robust fundamental matrix estimator with DEGENSAC~\cite{chum2005two}. We calculate the essential matrix from the estimated fundamental matrix using ground-truth intrinsics and then recover the poses using OpenCV~\cite{bradski2000opencv}. The tuning of hyper-parameters is performed on the validation subset of IMC-PT~\cite{jin2021image}, see the details in the supplementary material.

\begin{table}
\begin{center}
\resizebox{\columnwidth}{!}{
\begin{tabular}{lcccccc}
\toprule
\multirow{2}{*}{Methods} & \multirowcell{2}{Rotation \\ mAA ($10^\circ$)} & \multirowcell{2}{Translation \\ mAA ($10^\circ$)}\\
\\
\midrule
Shi-Tomasi~\cite{harris1988combined, shi1994good} + DISK~\cite{tyszkiewicz2020disk} & 0.744 & \textbf{\textcolor[HTML]{006795}{0.422}} \\
SIFT~\cite{lowe2004distinctive} + DISK~\cite{tyszkiewicz2020disk} & 0.7 & 0.387 \\
SuperPoint~\cite{detone2018superpoint} + DISK~\cite{tyszkiewicz2020disk} & 0.708 & 0.37 \\
R2D2~\cite{revaud2019r2d2} + DISK~\cite{tyszkiewicz2020disk} & \textbf{\textcolor[HTML]{006795}{0.751}} & 
0.403 \\
Key.Net~\cite{barroso2019key} + DISK~\cite{tyszkiewicz2020disk} & 0.666 & 0.327 \\
DISK \cite{tyszkiewicz2020disk} & \textbf{\textcolor[HTML]{B6321C}{0.813}} & \textbf{\textcolor[HTML]{B6321C}{{0.489}}} \\
REKD~\cite{lee2022self} + DISK~\cite{tyszkiewicz2020disk} & 0.601 & 0.271 \\
\midrule
NeSS-ST + DISK~\cite{tyszkiewicz2020disk} & \textbf{\textcolor[HTML]{3C8031}{0.767}} & \textbf{\textcolor[HTML]{3C8031}{0.438}} \\
\bottomrule
\end{tabular}
}
\end{center}
\caption{Evaluation on IMC-PT~\cite{jin2021image} with 2048 keypoints and full resolution images. We report mAA~\cite{yi2018learning, jin2021image} up to a 10 degrees threshold for rotation and translation. Best results are marked in \textbf{\textcolor[HTML]{B6321C}{red}}, 2nd best - in \textbf{\textcolor[HTML]{3C8031}{green}}, 3rd best - in \textbf{\textcolor[HTML]{006795}{blue}}.}
\label{tab:imcpt}
\end{table}

\paragraph{Results.}

Comparison in Table~\ref{tab:imcpt} shows that we considerably outperform other self-supervised approaches like SuperPoint~\cite{detone2018superpoint}, Key.Net~\cite{barroso2019key} and REKD~\cite{lee2022self}. We explain the gap between DISK~\cite{tyszkiewicz2020disk} and our method by the difference in the strategies that the methods employ to detect points. DISK~\cite{tyszkiewicz2020disk} tends to densely detect points on semantically meaningful objects (mostly, buildings), whereas our method doesn't employ any knowledge of scene semantics and, instead, relies on the local properties of points, which results in sparsified detections. Given that IMC-PT~\cite{jin2021image} contains a lot of extreme viewpoint and scale changes, the former strategy looks more advantageous in such an environment. Refer to the supplementary material for additional results.

\subsubsection{Evaluation on MegaDepth}

The MegaDepth~\cite{li2018megadepth} dataset is another photo-tourism collection of images that also provides depth maps and poses. As the IMC-PT dataset has a limited diversity of scenes and their semantics and has only 800 unique images, we create a test set from the MegaDepth dataset to perform the assessment with a larger diversity of data. In particular, our test set consists of 7.5k image pairs with 6k unique images sampled from 25 scenes belonging to 5 semantically different categories. This test set doesn't have any intersections neither with validation nor with test sequences of IMC-PT.

Since both IMC-PT and MegaDepth belong to the category of outdoors/photo-tourism datasets, we use the same descriptor, pose estimation routines and hyper-parameters for the evaluation.

\begin{table}
\begin{center}
\resizebox{\columnwidth}{!}{
\begin{tabular}{lcccccc}
\toprule
\multirow{2}{*}{Methods} & \multirowcell{2}{Rotation \\ mAA ($10^\circ$)} & \multirowcell{2}{Translation \\ mAA ($10^\circ$)}\\
\\
\midrule
Shi-Tomasi~\cite{harris1988combined, shi1994good} + DISK~\cite{tyszkiewicz2020disk} & 0.858 & 0.31 \\
SIFT~\cite{lowe2004distinctive} + DISK~\cite{tyszkiewicz2020disk} & 0.83 & 0.299 \\
SuperPoint~\cite{detone2018superpoint} + DISK~\cite{tyszkiewicz2020disk} & \textbf{\textcolor[HTML]{006795}{0.873}} & \textbf{\textcolor[HTML]{006795}{0.318}} \\
R2D2~\cite{revaud2019r2d2} + DISK~\cite{tyszkiewicz2020disk} & 0.871 & 0.312 \\
Key.Net~\cite{barroso2019key} + DISK~\cite{tyszkiewicz2020disk} & 0.84 & 0.278 \\
DISK~\cite{tyszkiewicz2020disk} & \textbf{\textcolor[HTML]{3C8031}{0.877}} & \textbf{\textcolor[HTML]{3C8031}{0.326}} \\
REKD~\cite{lee2022self} + DISK~\cite{tyszkiewicz2020disk} & 0.848 & 0.274 \\
\midrule
NeSS-ST + DISK~\cite{tyszkiewicz2020disk} & \textbf{\textcolor[HTML]{B6321C}{0.878}} & \textbf{\textcolor[HTML]{B6321C}{0.335}} \\
\bottomrule
\end{tabular}
}
\end{center}
\caption{Evaluation on MegaDepth~\cite{li2018megadepth} with 2048 keypoints and full resolution images. We report mAA~\cite{yi2018learning, jin2021image} up to a 10 degrees threshold for rotation and translation. Best results are marked in \textbf{\textcolor[HTML]{B6321C}{red}}, 2nd best - in \textbf{\textcolor[HTML]{3C8031}{green}}, 3rd best - in \textbf{\textcolor[HTML]{006795}{blue}}.}
\label{tab:megadepth}
\end{table}

\paragraph{Results.} Contents of Table~\ref{tab:megadepth} show that the evaluation on a more diverse set of data narrows the gap between methods with our method being marginally better than DISK~\cite{tyszkiewicz2020disk}. SuperPoint~\cite{detone2018superpoint} is the second-best method in the category of methods that don't employ reconstructed correspondence labels. Although our gains in rotation mAA compared to it are marginal, we obtain noticeably better translation estimates. Refer to the supplementary material for more results.

\subsubsection{Evaluation on ScanNet}

To assess the generalization ability of our method we perform the evaluation on the ScanNet~\cite{dai2017scannet} dataset that contains indoor video sequences provided with camera poses and depth maps. Following~\cite{ono2018lf, wang2020learning}, we create validation and test sets by sampling pairs from video sequences with different gaps between frames. Our test set consists of 21k image pairs with 39k unique images sampled from 100 test sequences of the dataset.

We perform the evaluation using the HardNet~\cite{mishchuk2017working} descriptor since on this dataset it shows performance that is superior to DISK~\cite{tyszkiewicz2020disk} for every method. Because ScanNet is captured on a single RGB camera, we employ a robust essential matrix estimator from OpenGV~\cite{kneip2014opengv} with the rest of the pipeline remaining the same as in previous evaluations. We use the validation subset of ScanNet for tuning hyper-parameters, see the details in the supplementary material.

\begin{table}
\begin{center}
\resizebox{\columnwidth}{!}{
\begin{tabular}{lcccccc}
\toprule
\multirow{2}{*}{Methods} & \multirowcell{2}{Rotation \\ mAA ($10^\circ$)} & \multirowcell{2}{Translation \\ mAA ($10^\circ$)}\\
\\
\midrule
Shi-Tomasi~\cite{harris1988combined, shi1994good} + HardNet~\cite{mishchuk2017working} & 0.568 & 0.196 \\
SIFT~\cite{lowe2004distinctive} + HardNet~\cite{mishchuk2017working} & 0.578 & 0.197 \\
SuperPoint~\cite{detone2018superpoint} + HardNet~\cite{mishchuk2017working} & \textbf{\textcolor[HTML]{006795}{0.611}} & \textbf{\textcolor[HTML]{006795}{0.217}} \\
R2D2~\cite{revaud2019r2d2} + HardNet~\cite{mishchuk2017working} & \textbf{\textcolor[HTML]{B6321C}{0.642}} & \textbf{\textcolor[HTML]{B6321C}{0.235}} \\
Key.Net~\cite{barroso2019key} + HardNet~\cite{mishchuk2017working} & 0.606 & 0.213 \\
DISK~\cite{tyszkiewicz2020disk} + HardNet~\cite{mishchuk2017working} & 0.528 & 0.162 \\
REKD~\cite{lee2022self} + HardNet~\cite{mishchuk2017working} & 0.464 & 0.142 \\
\midrule
NeSS-ST + HardNet~\cite{mishchuk2017working} & 
\textbf{\textcolor[HTML]{3C8031}{0.621}} & \textbf{\textcolor[HTML]{3C8031}{0.222}} \\
\bottomrule
\end{tabular}
}
\end{center}
\caption{Evaluation on ScanNet~\cite{dai2017scannet} with 2048 keypoints and full resolution images. We report mAA~\cite{yi2018learning, jin2021image} up to a 10 degrees threshold for rotation and translation. Best results are marked in \textbf{\textcolor[HTML]{B6321C}{red}}, 2nd best - in \textbf{\textcolor[HTML]{3C8031}{green}}, 3rd best - in \textbf{\textcolor[HTML]{006795}{blue}}.}
\label{tab:scannet}
\end{table}

\paragraph{Results.} Contrary to photo-tourism datasets that depict objects with a lot of texture, ScanNet~\cite{dai2017scannet} indoor environments contain a lot of surfaces with little to no texture. We believe that this is the main reason why DISK~\cite{tyszkiewicz2020disk} shows significantly inferior performance on this set of data (see Table~\ref{tab:scannet}). Our method, on the other hand, can consistently cope with the challenge and shows the second-best result achieving significant improvements compared to  SuperPoint~\cite{detone2018superpoint}, Key.Net~\cite{barroso2019key} and REKD~\cite{lee2022self}. R2D2~\cite{revaud2019r2d2} achieves the best results on this dataset; we found that the method is able to provide consistent matches on images with little texture and poor illumination conditions by performing detection along the contours of objects. Refer to the supplementary material for additional results.

To sum up, the experiments show that NeSS-ST is the only method in the top three across all the datasets (see Table~\ref{tab:overall_ranking}) as well as has the best performance among the self-supervised methods listed in Table~\ref{tab:self_supervised}. It has better generalization ability compared to most of state-of-the-art, compact model size and a fast running time (see Table~\ref{tab:model_comparison}).

\begin{table}
\begin{center}
\resizebox{\columnwidth}{!}{
\begin{tabular}{lccccc}
\toprule
 & HPatches & IMC-PT & MegaDepth & ScanNet \\
\midrule
First & Shi-Tomasi & DISK & \textbf{\textcolor[HTML]{3C8031}{NeSS-ST}} & R2D2 \\
Second & \textbf{\textcolor[HTML]{3C8031}{NeSS-ST}} & \textbf{\textcolor[HTML]{3C8031}{NeSS-ST}} & DISK & \textbf{\textcolor[HTML]{3C8031}{NeSS-ST}} \\
Third & SuperPoint & R2D2/Shi-Tomasi & SuperPoint & SuperPoint \\
\bottomrule
\end{tabular}
}
\end{center}
\caption{The top three ranking of detectors on downstream tasks.}
\label{tab:overall_ranking}
\end{table}

\begin{table*}
\begin{center}
\resizebox{\linewidth}{!}{
\begin{tabular}{lcccccc|c}
\toprule
& SuperPoint~\cite{detone2018superpoint} & R2D2~\cite{revaud2019r2d2} & Key.Net~\cite{barroso2019key} & DISK~\cite{tyszkiewicz2020disk} & REKD~\cite{lee2022self} & NeSS-ST & SS-ST \\
\midrule
Size (MB) & 4.96 & 1.85 & 0.02 & 4.17 & 99.12 & 3.54 & - \\
Inference time (ms) & 4.0 & 24.4 & 5.4 & 19.5 & 59.5 & 8.2 & \multirow{2}{*}{403.9} \\
Post-processing time (ms) & 1.7 & 2.2 & 2.2 & 0.7 & 2.0 & 7.4 &  \\
\bottomrule
\end{tabular}
}
\end{center}
\caption{Comparison of models sizes and running-time for $640 \times 480$ resolution images on NVIDIA 2080 Ti GPU. Please note that post-processing procedures (keypoint selection, sub-pixel localization) are not optimized.}
\label{tab:model_comparison}
\end{table*}

\subsection{Ablation study}

\subsubsection{Base detector ablation}
\label{sec:base_detector_abl}

Our method operates on top of the handcrafted Shi-Tomasi~\cite{harris1988combined, shi1994good} detector, however, NeSS can be used in combination with other detectors. The choice of a handcrafted detector over a learned detector is motivated by the size of patch $p$ that is required to maintain online ground truth generation (see Sec.~\ref{sec:kss}). Specifically, handcrafted detectors require a small neighbourhood to calculate the score compared to learned detectors that have a large receptive field. Another reason for giving the preference to handcrafted detectors is their good performance on downstream tasks, DoG, in particular, reported in~\cite{jin2021image}, which is also confirmed in our evaluation.

We pick the Shi-Tomasi~\cite{harris1988combined, shi1994good}  detector over the Harris~\cite{harris1988combined} detector since the keypoint selection criterion of the former better fits NeSS. More precisely, the Shi-Tomasi detector doesn't impose any restrictions on the shape of the second-moment matrix; an edge-like pattern and a weak blob-like pattern can look the same to the Shi-Tomasi detector. The Harris detector, on the other hand, draws a clear distinction between points and edges: eigenvalues of the second-moment matrix need to be relatively well-proportioned for a high Harris score.

We explore the performance of various handcrafted detectors by using them as base detectors for NeSS. Based on the summary given in Sec.~\ref{sec:related_work}, handcrafted detectors can be categorized into those that detect only spatial extrema, and those that also model the invariance to viewpoint changes. Since NeSS, by design, assesses the invariance of spatial extrema to viewpoint transformations (see Sec.~\ref{sec:kss}), we consider only the former kind of methods to serve as base detectors. In particular, we examine the determinant of the Hessian (DoH)~\cite{lindeberg1998feature} and the Laplacian of the Gaussian (LoG)~\cite{blostein1989multiscale, lindeberg1998feature} detectors. The latter detector presents a particular interest since DoG is the approximation of scale-normalized LoG~\cite{lindeberg1998feature, lowe2004distinctive}. All base detectors in the evaluation use sub-pixel localization.

\paragraph{Results.} Table~\ref{tab:base_detector_abl} shows that NeSS noticeably improves the performance of all base detectors; however,  at the same time, the results indicate that the gains provided by NeSS depend on the performance of an employed base detector. Shi-Tomasi~\cite{harris1988combined, shi1994good} has the best performance among all base detectors and demonstrates the best results when combined with NeSS. Interestingly, Shi-Tomasi shows noticeable improvements over Harris on translation estimation that correlates well with the theoretical justification behind the design of the Shi-Tomasi detector~\cite{shi1994good}. More results can be found in the supplementary material.

\begin{table}
\begin{center}
\begin{tabular}{lcccccc}
\toprule
\multirow{2}{*}{Method} & \multirowcell{2}{Rotation \\ mAA ($10^\circ$)} & \multirowcell{2}{Translation \\ mAA ($10^\circ$)}\\
\\
\midrule
Shi-Tomasi~\cite{harris1988combined, shi1994good} & \textbf{\textcolor[HTML]{B6321C}{0.744}} & \textbf{\textcolor[HTML]{B6321C}{0.422}} \\
Harris~\cite{harris1988combined} & 0.743 & 0.405 \\
DoH~\cite{lindeberg1998feature} & 0.694 & 0.363 \\
LoG~\cite{blostein1989multiscale, lindeberg1998feature} & 0.708 & 0.376 \\

\midrule
NeSS-ST & \textbf{\textcolor[HTML]{B6321C}{0.766}} & \textbf{\textcolor[HTML]{B6321C}{0.438}} \\
NeSS-DoH & 0.742 & 0.39 \\
NeSS-LoG & 0.763 & 0.421 \\
\bottomrule
\end{tabular}
\end{center}
\caption{Base detector ablation on IMC-PT~\cite{jin2021image} with DISK~\cite{tyszkiewicz2020disk} descriptor, 2048 keypoints and full resolution images. We report mAA~\cite{yi2018learning, jin2021image} up to a 10 degrees threshold for rotation and translation. Best results are marked in \textbf{\textcolor[HTML]{B6321C}{red}}.}
\label{tab:base_detector_abl}
\end{table}

\subsubsection{Stability score design ablation}
\label{sec:stability_score_abl}

We highlight the importance of accounting for the uncertainty of a keypoint location in the stability score (see Eq.~\ref{eq:sigma} and Eq.~\ref{eq:lambda}) by conducting experiments with another score. In particular, using the notation from Eq.~\ref{eq:sigma}, we formulate the {\it repeatability score (RS)} that acts like $\epsilon$-pixel repeatability measure:

\begin{equation}
\label{eq:r}
\mathbf{r} = \sum^m_{j=1} \frac{\mathds{1}^{\mathit{rep}}(\mathcal{H}^{-1}_j\hat{\mathbf{l}}_j - \mathbf{k}, \epsilon)}{m}.
\end{equation}

We define $\mathds{1}^{\mathit{rep}}(\mathcal{H}^{-1}_j\hat{\mathbf{l}}_j - \mathbf{k}, \epsilon)$ as an indicator function that gives $1$ if $|\mathcal{H}^{-1}_j\hat{\mathbf{l}}_j - \mathbf{k}|_\infty < \epsilon$ and $0$ otherwise. Setting $\epsilon=1$, we train the neural network to predict the {\it neural repeatability score (NeRS)} in the similar to Eq.~\ref{eq:ss_loss} manner.

To emphasize the role of the neural network in our design we build detectors that operate without it by directly calculating SS and RS for all potential feature points on an image. Apart from the computational concerns, such design choice requires picking a threshold to filter the noise (see Sec.~\ref{sec:training}). This poses a problem since the optimal threshold depends on the intensity of depicted patterns that varies from image to image. Utilizing filtering during the training is exempt from this limitation: on the contrary, it allows to select only reliable points for learning a new rule from the data and, hence, to a certain extent, overcome the limitations of homography sampling. Refer to the supplementary material for the details.

\begin{table}
\begin{center}
\begin{tabular}{lcccccc}
\toprule
\multirow{2}{*}{Method} & \multirowcell{2}{Rotation \\ mAA ($10^\circ$)} & \multirowcell{2}{Translation \\ mAA ($10^\circ$)}\\
\\
\midrule
SS-ST & 0.757 & 0.417 \\
RS-ST & 0.622 & 0.273 \\
NeSS-ST & \textbf{\textcolor[HTML]{B6321C}{0.766}} & \textbf{\textcolor[HTML]{B6321C}{0.438}} \\
NeRS-ST & 0.75 & 0.41 \\
\bottomrule
\end{tabular}
\end{center}
\caption{Stability score design ablation on IMC-PT~\cite{jin2021image} with DISK~\cite{tyszkiewicz2020disk} descriptor, 2048 keypoints and full resolution images. We report mAA~\cite{yi2018learning, jin2021image} up to a 10 degrees threshold for rotation and translation. Best results are marked in \textbf{\textcolor[HTML]{B6321C}{red}}.}
\label{tab:ss_abl}
\end{table}

\begin{table}
\begin{center}
\begin{tabular}{lcccccc}
\toprule
\multirow{2}{*}{Method} & \multirowcell{2}{Repeatability \\ (3 px)} & \multirowcell{2}{MMA \\ (3 px)}\\
\\
\midrule
SS-ST & 0.339 & 0.655 \\
RS-ST & \textbf{\textcolor[HTML]{B6321C}{0.423}} & \textbf{\textcolor[HTML]{B6321C}{0.668}} \\
\bottomrule
\end{tabular}
\end{center}
\caption{Stability score design ablation on HPatches~\cite{balntas2017hpatches} with DISK~\cite{tyszkiewicz2020disk} descriptor, 2048 keypoints and full resolution images. We report MMA~\cite{mikolajczyk2005performance, dusmanu2019d2} and repeatability~\cite{mikolajczyk2005comparison} under a 3-pixel threshold. Best results are marked in \textbf{\textcolor[HTML]{B6321C}{red}}.}
\label{tab:ss_hpatches_abl}
\end{table}

\paragraph{Results.} Results presented in Table~\ref{tab:ss_abl} indicate that the assessment of the local uncertainty of a keypoint location (SS, NeSS) yields a better criterion compared to repeatability (RS, NeRS) when the downstream task of pose estimation is considered. Notably, although the RS criterion provides keypoints for training that have better repeatability and MMA compared to SS (see Table~\ref{tab:ss_hpatches_abl}), the model trained using the latter criterion shows much better results on the downstream task (see Table~\ref{tab:ss_abl}). Table~\ref{tab:ss_abl} and Table~\ref{tab:model_comparison} show the benefits of utilizing the neural network (NeSS, NeRS) over the algorithmic approach (SS, RS) from both performance and computational standpoints. More results can be found in the supplementary material.

\section{Conclusion}
In this work, we proposed the NeSS-ST detector that combines the handcrafted Shi-Tomasi detector and the neural stability score. The method doesn't require any reconstructed correspondence labels and can be trained from arbitrary sets of images without the need for dataset pre-labeling. It achieves state-of-the-art performance on a variety of datasets and downstream tasks, has good generalization and consistently outperforms other self-supervised methods. In the future, we plan to address the main limitation of our method which is the lack of illumination invariance as well as use inferences from the evaluation of methods like DISK and R2D2 to improve our keypoint detection strategy. Additionally, NeSS may be used as a weight in non-linear pose refinement or metric learning of feature descriptors.



{\small
\bibliographystyle{iccv2023AuthorKit/ieee_fullname}
\bibliography{bib/refs}
}

\maketitlesupplementary

\section{Additional method details}

\subsection{Stability score}

\begin{figure*}
\begin{center}
\includegraphics[align=c, width=\linewidth]{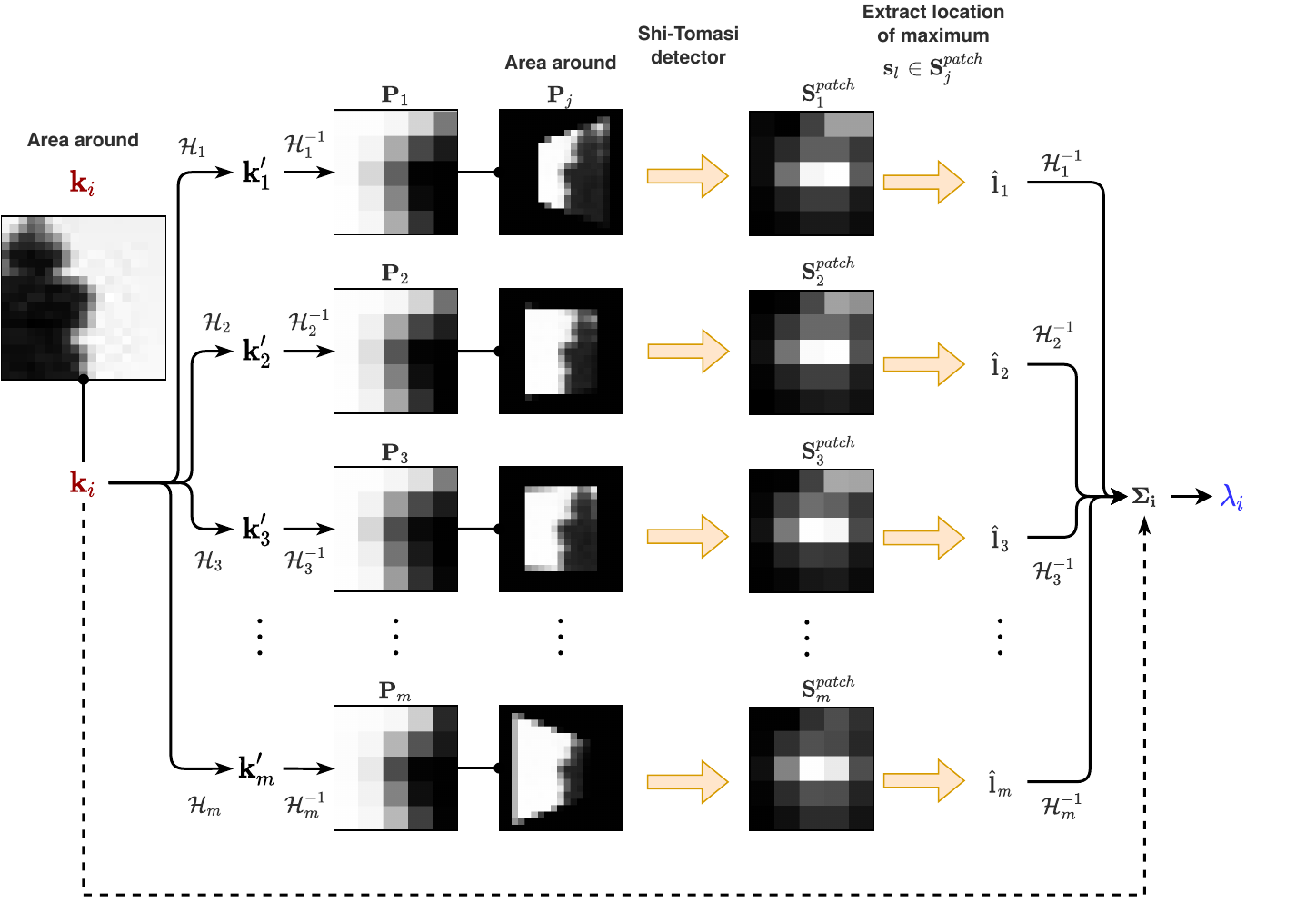}
\end{center}
   \caption{For each selected point $\mathbf{k}_i$ we calculate the ground truth stability score $\mathbf{\lambda}_i$. Firstly, we generate a set of deformed patches $\{\mathbf{P}_j\}^m_{j=1}$ and run the Shi-Tomasi detector on patches to obtain a set of score patches $\{\mathbf{S}^{\mathit{patch}}_j\}^{m}_{j=1}$. For each $\mathbf{S}^{\mathit{patch}}_j$ we extract the location of its maximum score $\hat{\mathbf{l}}_j$ getting a set $\{\hat{\mathbf{l}}_j\}^m_{j=1}$. By transforming the elements of the set with $\{\mathcal{H}^{-1}_j\}^{m}_{j=1}$,  we estimate $\mathbf{\Sigma}_i$ and calculate $\mathbf{\lambda}_i$.}
\label{fig:ground_truth}
\end{figure*}

The process of ground truth generation is illustrated in Fig.~\ref{fig:ground_truth}. The restriction of deformations along the x and y axes, which is mentioned in the main text, is controlled by a parameter $d$. The parameter describes the length of an edge of a square embedded at the center of a larger square of size $pd$ (see Sec.~3.2 of the main text). When deformations along x and y are sampled, they cannot distort the larger square in a way that its corners end up inside the square described by the parameter $d$. See the implementation in the GitHub repository for more details; file {\it source/projective/homography.py}, function {\it sample\_homography}, parameter {\it scale\_factor}.

\subsection{Neural stability score}

We explore the influence of parameters $t_\mathit{Shi}$ and $d$ on the performance of NeSS-ST. We train the network with different $t_\mathit{Shi}$ and  $d$ and test obtained models on the validation subset of IMC-PT~\cite{jin2021image}.

\begin{figure*}
\begin{center}
\includegraphics[width=0.95\linewidth]{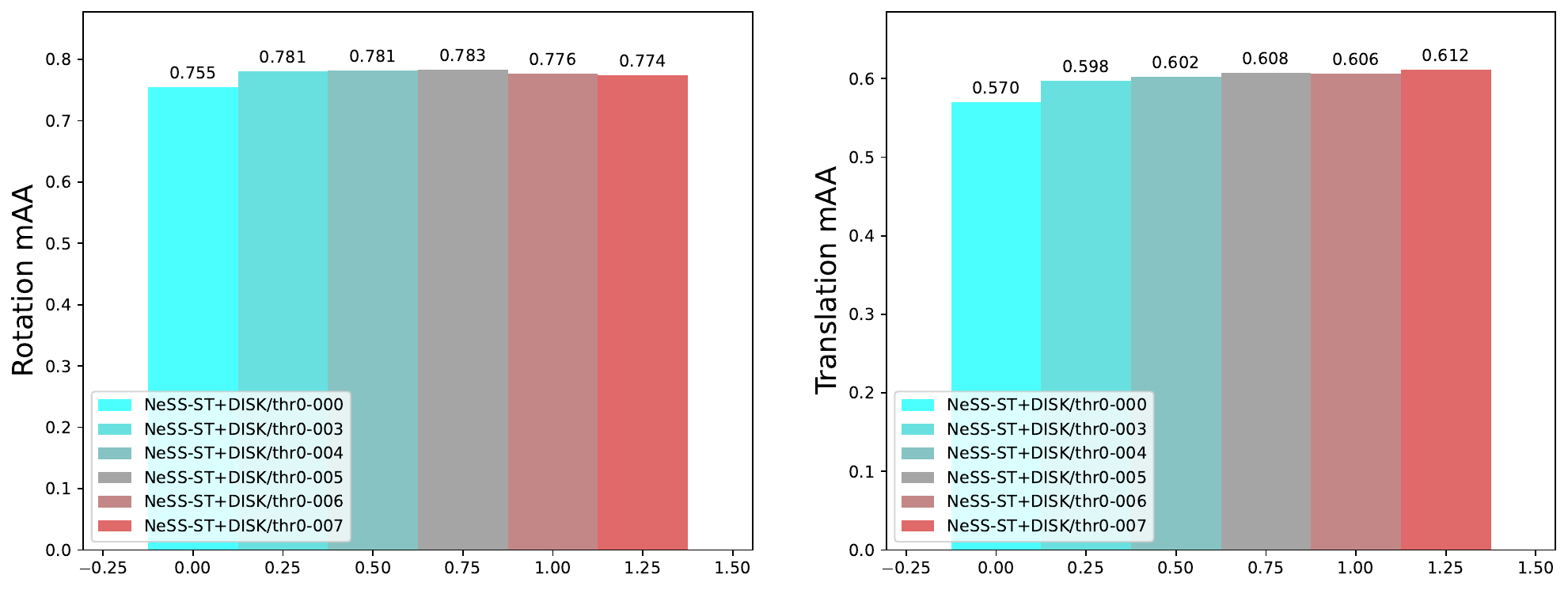}
\end{center}
\caption{NeSS-ST models trained with different values of $t_\mathit{Shi}$ (0.0, 0.003, 0.005, 0.006 and 0.007) evaluated on the validation set of IMC-PT~\cite{jin2021image} with 2048 keypoints and full resolution images. We report mAA~\cite{yi2018learning, jin2021image} up to a 10 degrees threshold for rotation and translation.}
\label{fig:nessst_filtering}
\end{figure*}

\begin{figure*}
\begin{center}
\includegraphics[width=0.95\linewidth]{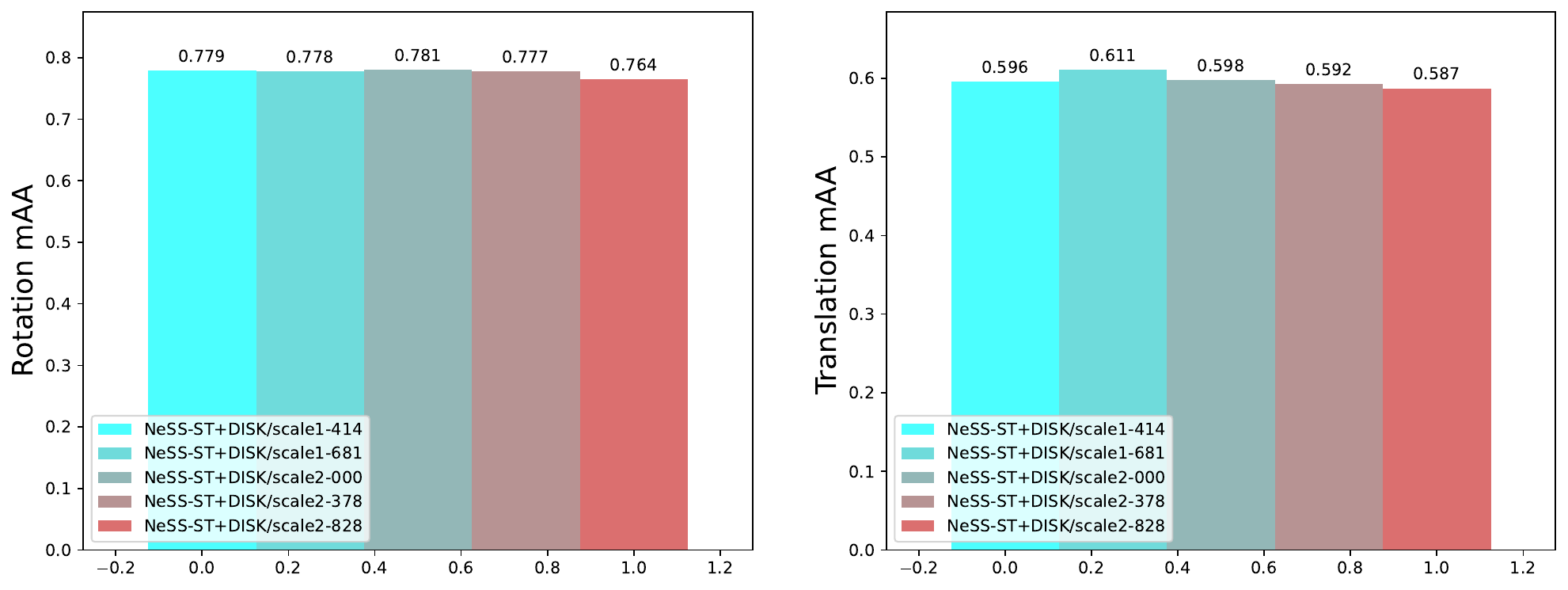}
\end{center}
\caption{NeSS-ST models trained with different values of $d$ (1.414, 1.681, 2.0, 2.378 and 2.828) evaluated on the validation set of IMC-PT~\cite{jin2021image} with 2048 keypoints and full resolution images. We report mAA~\cite{yi2018learning, jin2021image} up to a 10 degrees threshold for rotation and translation.}
\label{fig:nessst_homography}
\end{figure*}

\begin{figure}
\begin{center}
\includegraphics[width=0.95\linewidth]{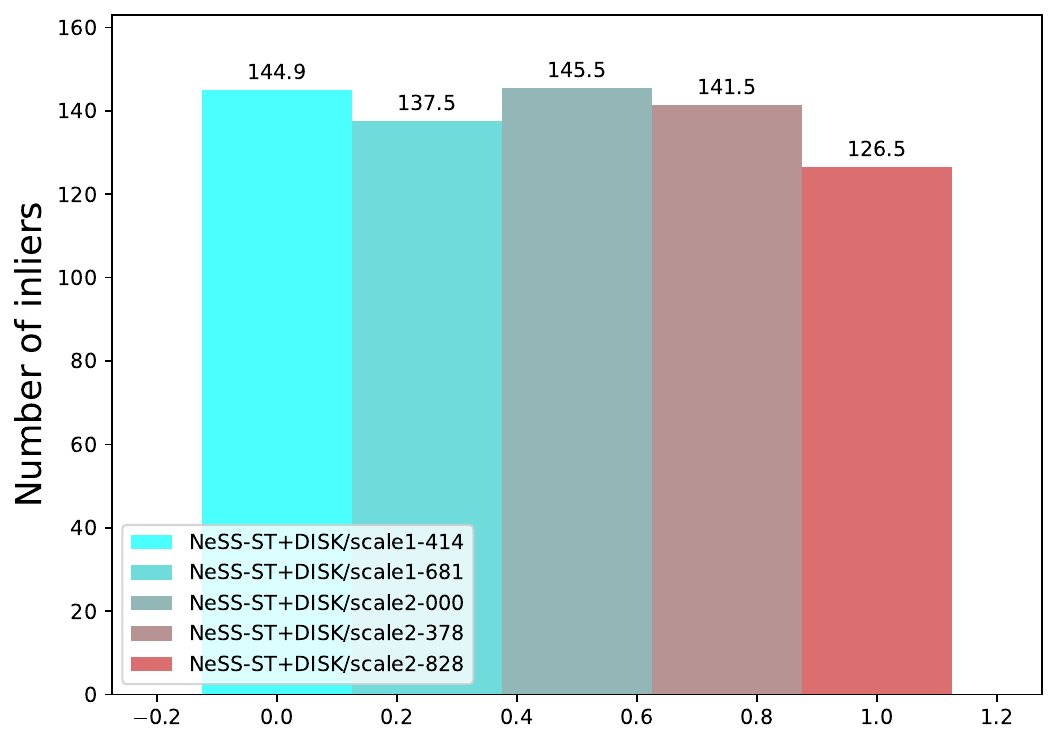}
\end{center}
\caption{NeSS-ST models trained with different values of $d$ (1.414, 1.681, 2.0, 2.378 and 2.828) evaluated on the validation set of IMC-PT~\cite{jin2021image} with 2048 keypoints and full resolution images. We report the number of inliers.}
\label{fig:nessst_homography_inliers}
\end{figure}

Based on the results of Fig.~\ref{fig:nessst_filtering}, Fig.~\ref{fig:nessst_homography} and Fig.~\ref{fig:nessst_homography_inliers}, we pick $t_\mathit{Shi}=0.005$ and $d=2.0$.

\section{Additional experiments}

During experiments, non-maximum suppression sizes, score thresholds and parameters that are related to the structure of the model, \eg the number of scales in Key.Net~\cite{barroso2019key}, are set for each model according to the values provided by the authors.

\subsection{Evaluation on HPatches}

\begin{figure*}
\begin{center}
\begin{subfigure}[b]{0.49\linewidth}
\begin{center}
\includegraphics[width=\linewidth]{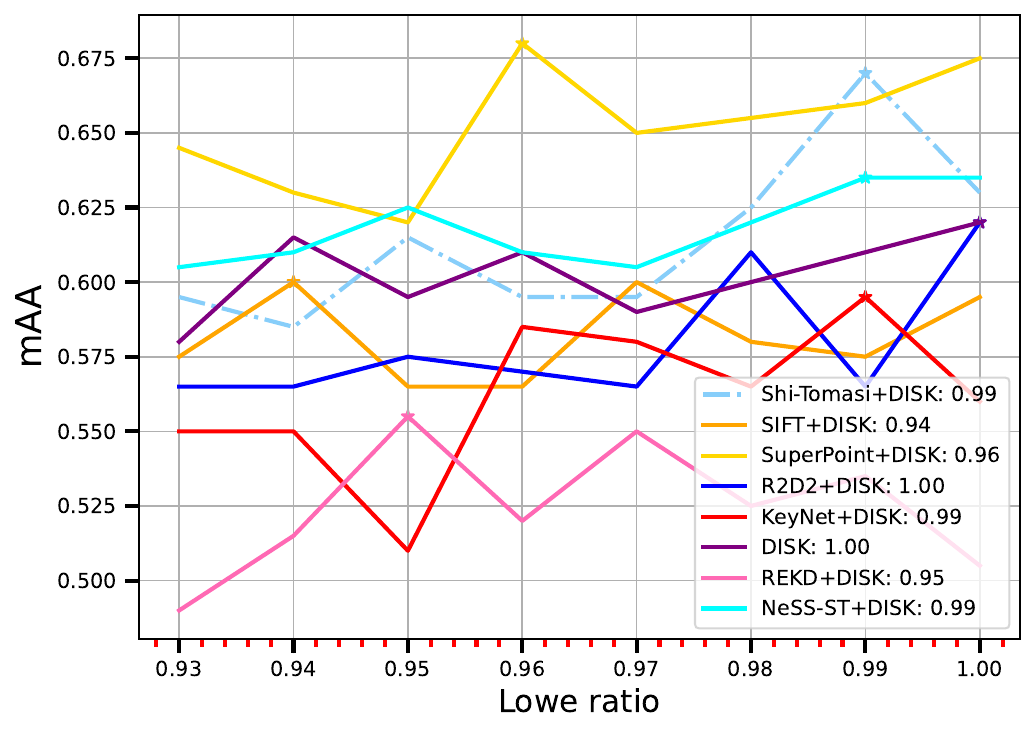}
  \caption{Lowe ratio tuning.}
\end{center}
\end{subfigure}
\begin{subfigure}[b]{0.49\linewidth}
\begin{center}
\includegraphics[width=\linewidth]{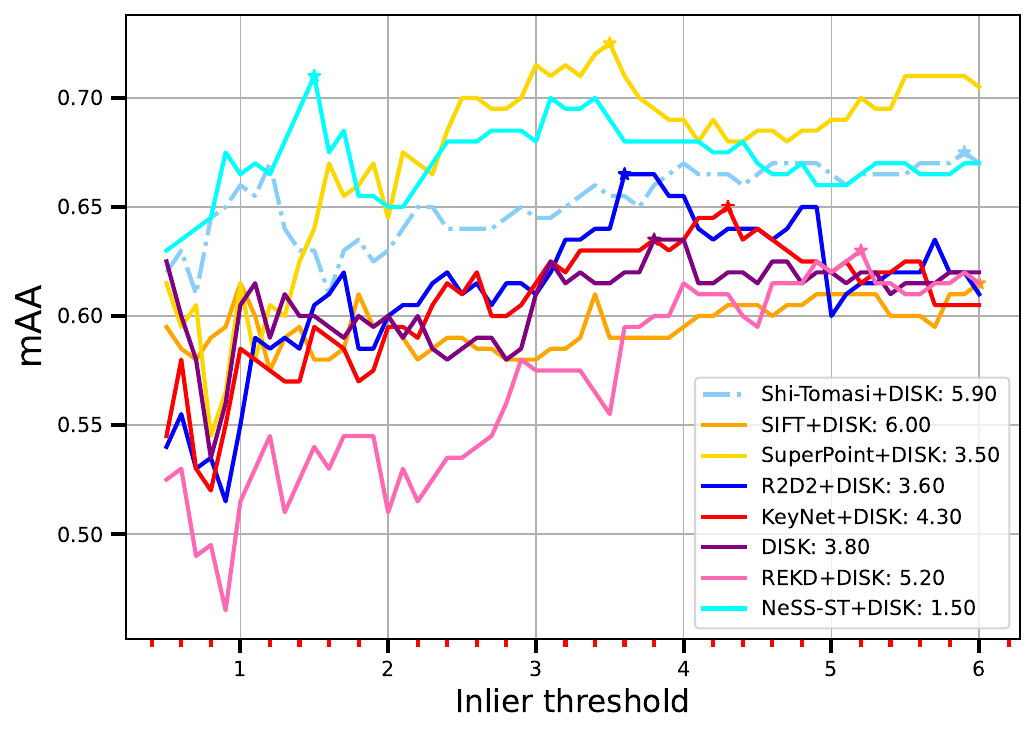}
\caption{Inlier threshold tuning.}
\end{center}
\end{subfigure}
\caption{Hyper-parameter tuning on the validation set of HPatches~\cite{balntas2017hpatches} with with 2048 keypoints and full resolution images. We report homography estimation mAA~\cite{detone2018superpoint, wang2020learning, yi2018learning, jin2021image} up to a 5-pixel threshold for different hyper-parameter values.}
\label{fig:hpatches_tuning}
\end{center}
\end{figure*}

We report repeatability~\cite{mikolajczyk2005comparison} under different pixel thresholds following~\cite{detone2018superpoint}. We  perform hyper-parameter fine-tuning for the homography estimation task~\cite{detone2018superpoint, wang2020learning}. We use images left out from the test set~\cite{dusmanu2019d2} as a validation set to get a general picture of the dependency between mAA and hyper-parameters (see Fig.~\ref{fig:hpatches_tuning}). Because the validation set of HPatches consists only of 48 unique images and 40 image pairs, using the best parameters obtained on it during the evaluation on the test set gives inadequate results. To overcome the problem of a small validation set size we firstly take Lowe ratio parameters fine-tuned on IMC-PT~\cite{jin2021image} (see Table~\ref{tab:imc_pt_tuning}). We reason it is a good approximation since IMC-PT also has strong viewpoint changes like HPatches. Next, we notice that for some models increasing the inlier threshold up to a limit leads to overall improvements, hence, we select a high inlier threshold for all models. This decision is based on the fact that for other datasets inlier thresholds are not far off from each other (see Table~\ref{tab:imc_pt_tuning} and Table~\ref{tab:scannet_tuning}). We estimate homographies using an OpenCV~\cite{bradski2000opencv} routine with 10000 iterations and a 0.9999 confidence level.

\begin{figure*}
\begin{center}
\includegraphics[width=0.98\linewidth]{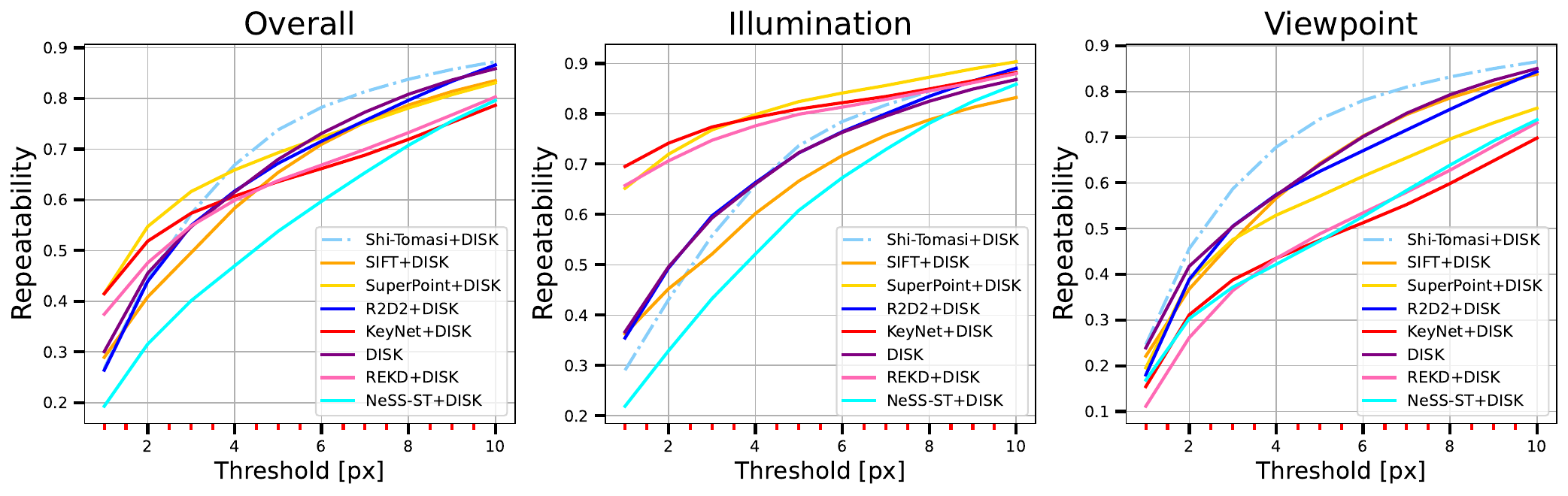}
\end{center}
\caption{Evaluation on HPatches~\cite{balntas2017hpatches} with 2048 keypoints and full resolution images. We report repeatability~\cite{mikolajczyk2005comparison}.}
\label{fig:hpatches_repeatability}
\end{figure*}

Fig.~\ref{fig:hpatches_repeatability} shows that our method has much lower repeatability compared to Shi-Tomasi~\cite{harris1988combined, shi1994good}, however, keeps on par with it on the homography estimation task.
Hyper-parameters reported in Table~\ref{tab:hpatches_tuning} provide significantly better results for all models among those that we tried on the test set. Although this kind of fine-tuning procedure is less principled than the one that we employ for IMC-PT~\cite{jin2021image} and ScanNet~\cite{dai2017scannet}, we believe that it is the best option available given that HPatches doesn't have enough data to allow a proper hyper-parameter tuning.

\begin{table}
\begin{center}
\resizebox{\columnwidth}{!}{
\begin{tabular}{lcccccc}
\toprule
\multirow{2}{*}{Methods} & Lowe ratio & Inlier threshold\\
\\
\midrule
Shi-Tomasi~\cite{harris1988combined, shi1994good} + DISK~\cite{tyszkiewicz2020disk} & 0.99 & 5.9 \\
SIFT~\cite{lowe2004distinctive} + DISK~\cite{tyszkiewicz2020disk} & 0.98 & 6.0 \\
SuperPoint~\cite{detone2018superpoint} + DISK~\cite{tyszkiewicz2020disk} & 0.98 & 5.4 \\
R2D2~\cite{revaud2019r2d2} + DISK~\cite{tyszkiewicz2020disk} & 0.98 & 5.4 \\
Key.Net~\cite{barroso2019key} + DISK~\cite{tyszkiewicz2020disk} & 0.99 & 5.9 \\
DISK~\cite{tyszkiewicz2020disk} & 0.98 & 5.4 \\
REKD~\cite{lee2022self} + DISK~\cite{tyszkiewicz2020disk} & 0.98 & 5.2 \\
\midrule
NeSS-ST + DISK~\cite{tyszkiewicz2020disk} & 0.99 & 5.4 \\
\bottomrule
\end{tabular}
}
\end{center}
\caption{Hyper-parameters used for homography estimation on HPatches~\cite{balntas2017hpatches}.}
\label{tab:hpatches_tuning}
\end{table}

\subsection{Evaluation on downstream tasks}

\subsubsection{Evaluation on IMC-PT}
\label{sec:imc_pt}

We provide accuracy-threshold plots for the evaluation on a downstream task of relative pose estimation on IMC-PT~\cite{jin2021image} from which we calculate mAA~\cite{yi2018learning, jin2021image}. We use the validation subset of IMC-PT for hyper-parameter tuning. The set consists of 274 unique images that form 11.5k image pairs. In accordance with the protocol from~\cite{jin2021image}, we first fine-tune the Lowe ratio, then - the inlier threshold. We plot mAA curves for different values of hyper-parameters for both rotation and translation. The best parameter is selected according to the largest sum of mAA for rotation and translation. We estimate the fundamental matrix by employing a robust estimator with DEGENSAC~\cite{chum2005two} using 200000 iterations and a 0.9999 confidence level. Additionally, we conduct experiments with different numbers of keypoints: we pick regimes of 128 and 512 keypoints to assess the ability of detectors to operate with a limited number of keypoints.

\begin{figure*}
\begin{center}
\begin{subfigure}[b]{0.91\linewidth}
\begin{center}
\includegraphics[width=\linewidth]{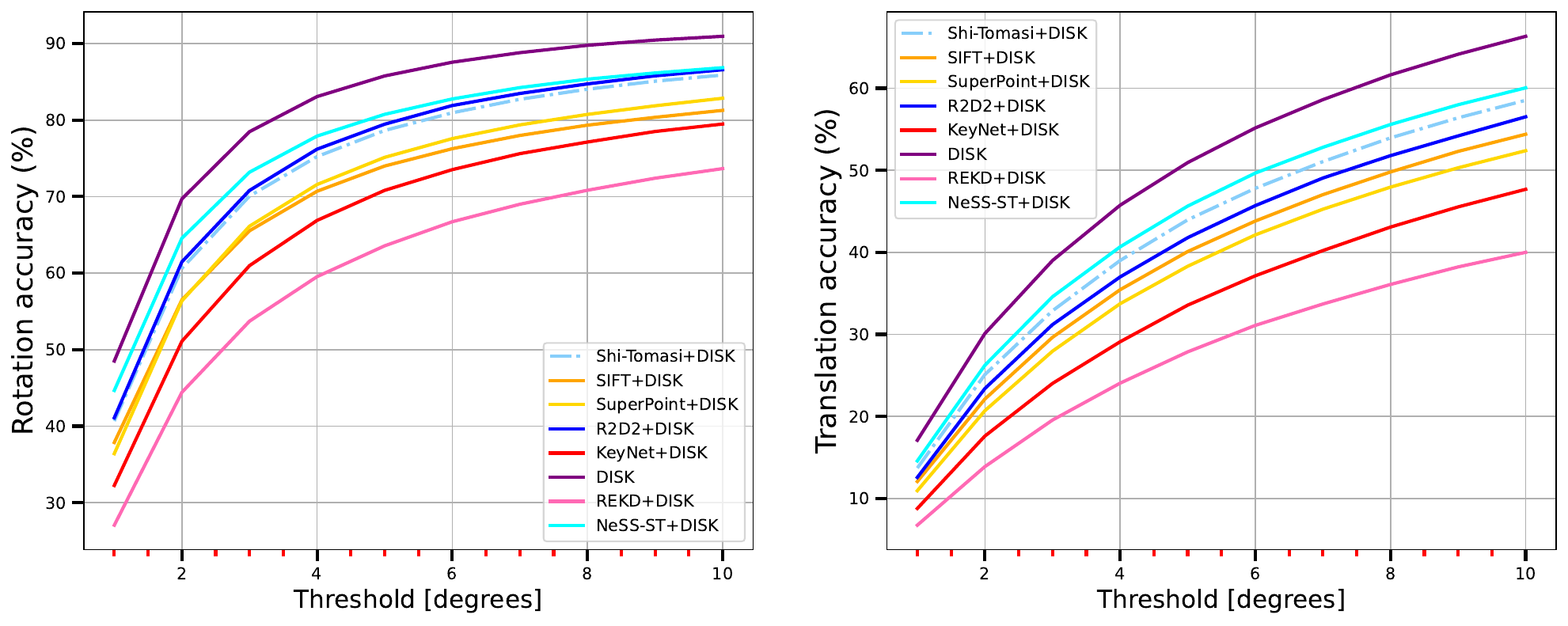}
\caption{Evaluation on IMC-PT~\cite{balntas2017hpatches}.}
\label{fig:imcpt}
\end{center}
\end{subfigure}
\begin{subfigure}[b]{0.91\linewidth}
\begin{center}
\includegraphics[width=\linewidth]{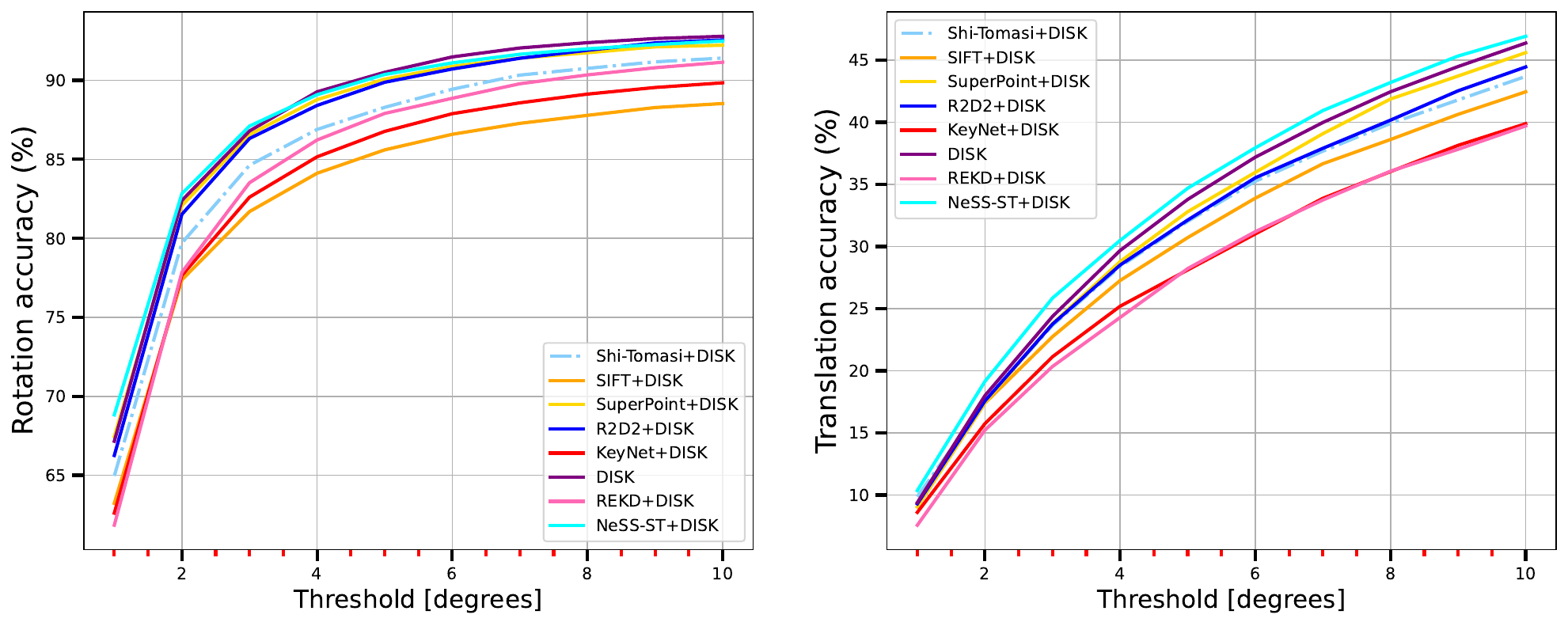}
\caption{Evaluation on MegaDepth~\cite{li2018megadepth}.}
\label{fig:megadepth}
\end{center}
\end{subfigure}
\begin{subfigure}[b]{0.91\linewidth}
\begin{center}
\includegraphics[width=\linewidth]{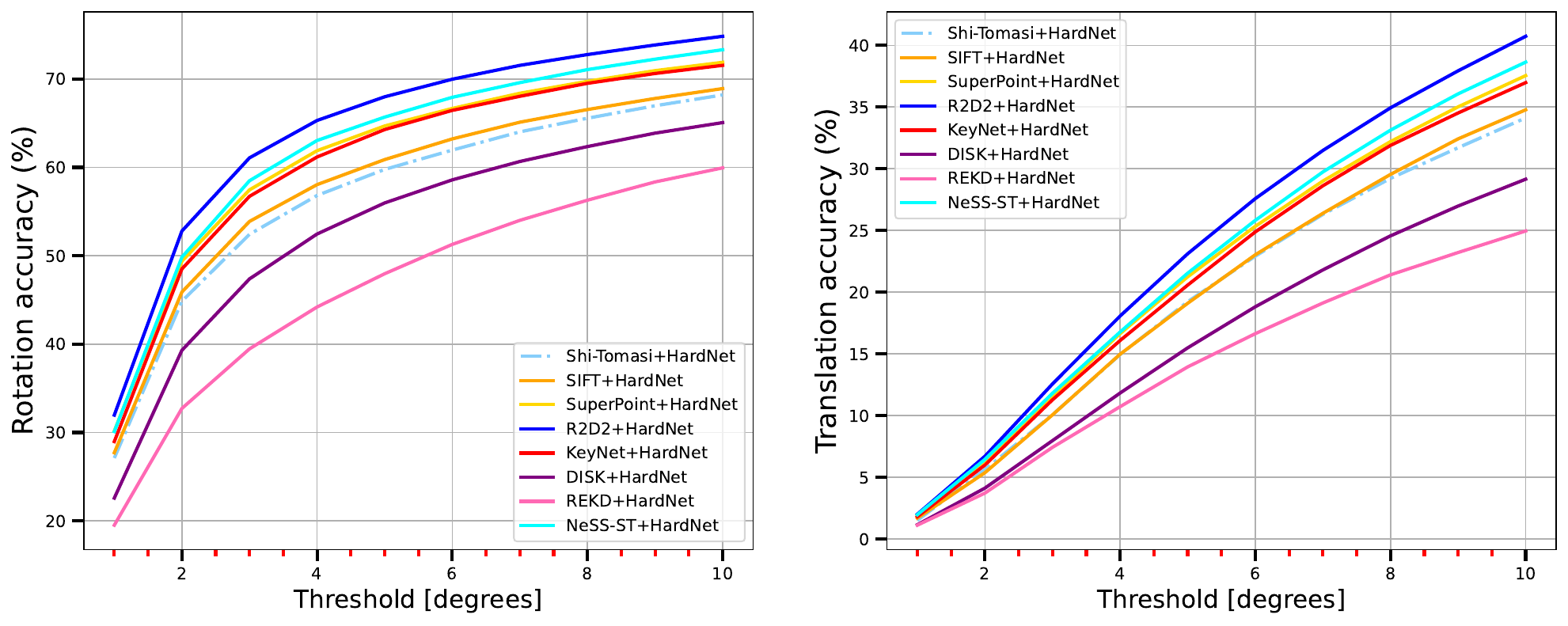}
\caption{Evaluation on ScanNet~\cite{dai2017scannet}.}
\label{fig:scannet}
\end{center}
\end{subfigure}
\caption{Evaluation on the downstream task of pose estimation with 2048 keypoints and full resolution images. We report pose estimation accuracy~\cite{yi2018learning, wang2020learning, jin2021image} in \%.}
\end{center}
\end{figure*}

\begin{figure*}
\begin{center}
\begin{subfigure}[b]{0.98\linewidth}
\begin{center}
\includegraphics[width=\linewidth]{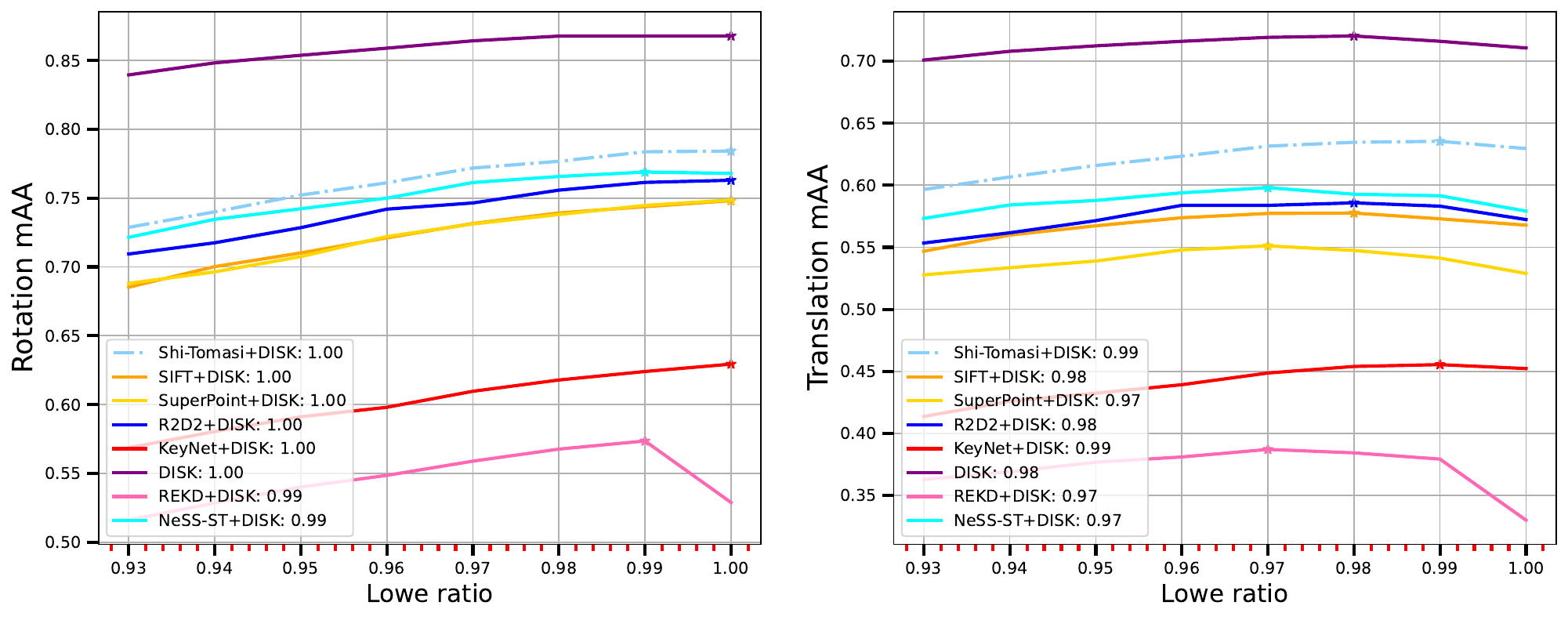}
  \caption{Lowe ratio tuning.}
\end{center}
\end{subfigure}
\begin{subfigure}[b]{0.98\linewidth}
\begin{center}
\includegraphics[width=\linewidth]{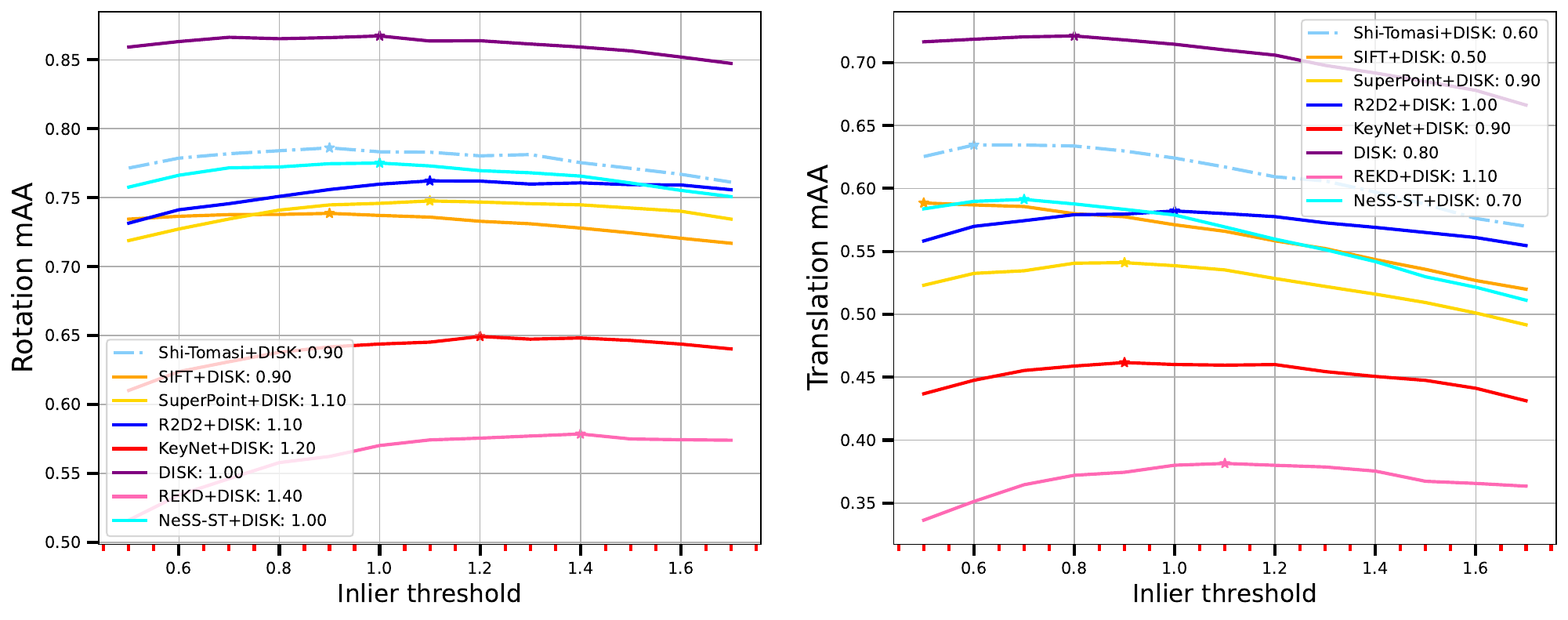}
\caption{Inlier threshold tuning.}
\end{center}
\end{subfigure}
\caption{Hyper-parameter tuning on the validation set of IMC-PT~\cite{jin2021image} with with 2048 keypoints and full resolution images. We report mAA~\cite{yi2018learning, jin2021image} up to a 10 degrees threshold for rotation and translation for different hyper-parameter values.}
\label{fig:imc_pt_tuning}
\end{center}
\end{figure*}

\begin{figure*}
\begin{center}
\begin{subfigure}[b]{0.98\linewidth}
\begin{center}
\includegraphics[width=0.98\linewidth]{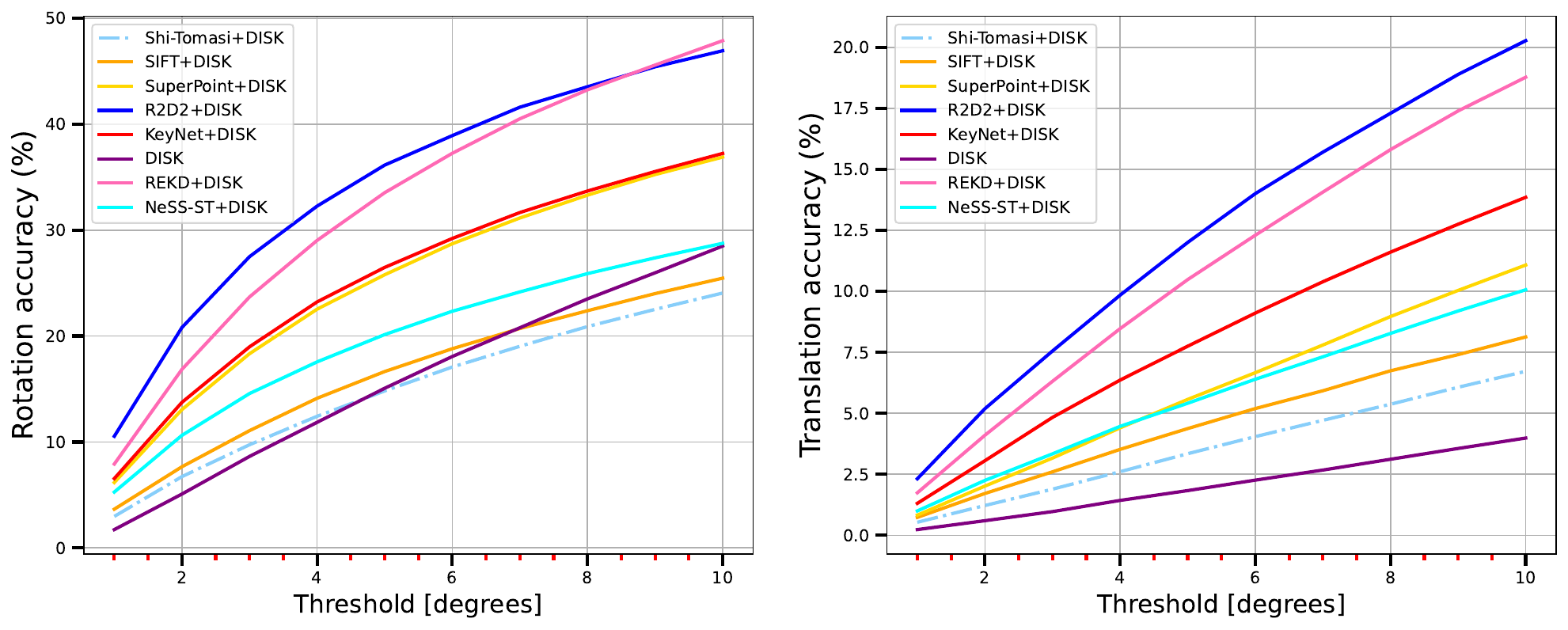}
\caption{128 keypoints.}
\label{fig:imcpt_128}
\end{center}
\end{subfigure}
\begin{subfigure}[b]{0.98\linewidth}
\begin{center}
\includegraphics[width=0.98\linewidth]{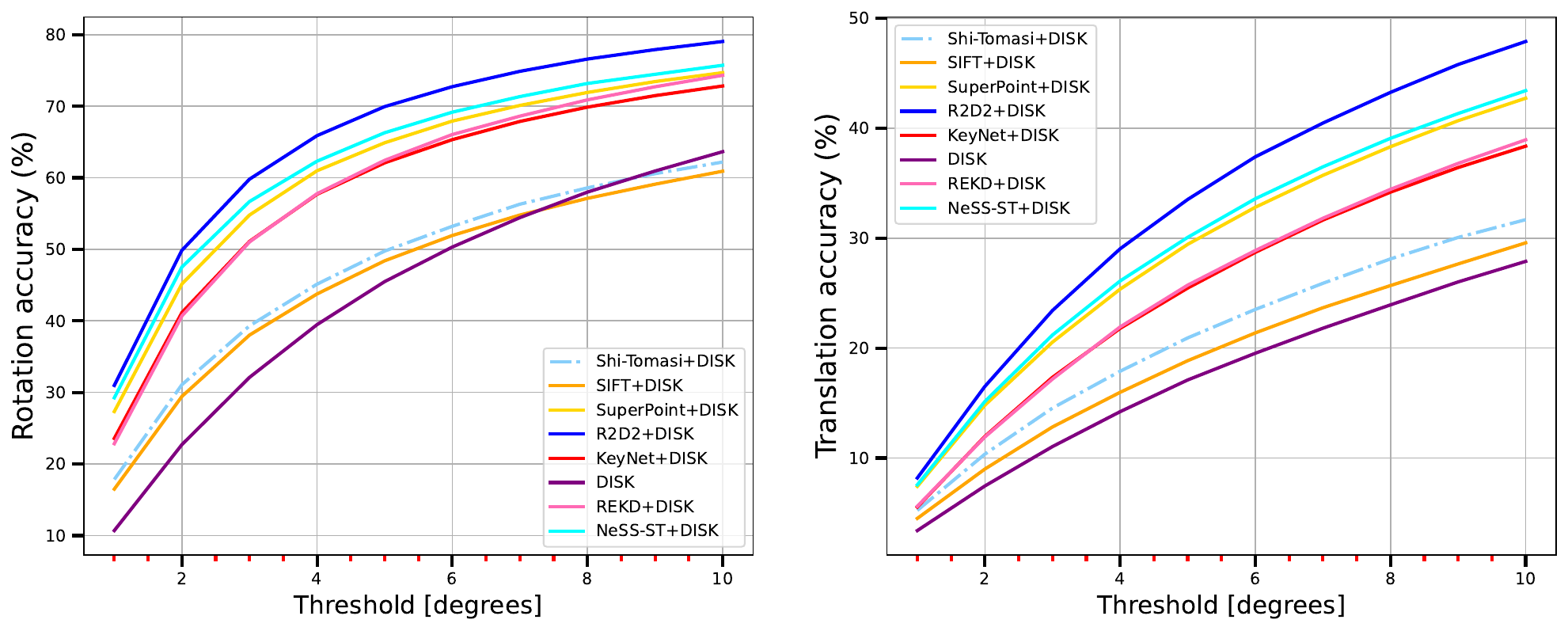}
\caption{512 keypoints.}
\label{fig:imcpt_512}
\end{center}
\end{subfigure}
\caption{Ablation on IMC-PT~\cite{balntas2017hpatches} with different numbers of keypoints and full resolution images. We report pose estimation accuracy~\cite{yi2018learning, wang2020learning, jin2021image} in \%.}
\end{center}
\end{figure*}

Fig.~\ref{fig:imcpt} shows that NeSS-ST consistently outperforms other self-supervised approaches over all thresholds. Fig.~\ref{fig:imc_pt_tuning} illustrates the dependency between the hyper-parameters and mAA. We report hyper-parameters selected for each model. We found that slightly changing hyper-parameters for some models improves their results on the test set, hence, the parameters used during the evaluation are slightly different from those selected on the validation set, see Table~\ref{tab:imc_pt_tuning}. Fig.~\ref{fig:imcpt_128} and Fig.~\ref{fig:imcpt_512} show that our method doesn't deal well with the decreased number of points. Firstly, we believe that it is related to our training setup where we select 1024 points per image. Secondly, in a few-point scenario points with higher repeatability, which our method doesn't look for, appear to present a better choice. We believe that this flaw in our method can be remedied by adding a term to the loss function that encourages correct predictions for different numbers of points.

\begin{table}
\begin{center}
\resizebox{\columnwidth}{!}{
\begin{tabular}{lcccccc}
\toprule
\multirow{2}{*}{Methods} & Lowe ratio & Inlier threshold\\
\\
\midrule
Shi-Tomasi~\cite{harris1988combined, shi1994good} + DISK~\cite{tyszkiewicz2020disk} & 0.99 & 0.6 \\
SIFT~\cite{lowe2004distinctive} + DISK~\cite{tyszkiewicz2020disk} & 0.98 & 0.6 \\
SuperPoint~\cite{detone2018superpoint} + DISK~\cite{tyszkiewicz2020disk} & 0.98 & 1.0 \\
R2D2~\cite{revaud2019r2d2} + DISK~\cite{tyszkiewicz2020disk} & 0.99 & 1.1 \\
Key.Net~\cite{barroso2019key} + DISK~\cite{tyszkiewicz2020disk} & 0.99 & 1.1 \\
DISK~\cite{tyszkiewicz2020disk} & 0.98 & 0.7 \\
REKD~\cite{lee2022self} + DISK~\cite{tyszkiewicz2020disk} & 0.98 & 1.1(1.3) \\
\midrule
NeSS-ST + DISK~\cite{tyszkiewicz2020disk} & 0.99 & 0.6 \\
\bottomrule
\end{tabular}
}
\end{center}
\caption{
Hyper-parameters used for fundamental matrix estimation on IMC-PT~\cite{jin2021image}. Tuned hyper-parameters, if different, are provided in brackets.}
\label{tab:imc_pt_tuning}
\end{table}

\subsubsection{Evaluation on MegaDepth}

We provide accuracy-threshold plots for the evaluation on the downstream task of relative pose estimation on MegaDepth~\cite{li2018megadepth} from which we calculate mAA~\cite{yi2018learning, jin2021image}.

Fig.~\ref{fig:megadepth} shows that NeSS-ST consistently outperforms all methods on translation estimation accuracy and is in the top three on rotation estimation accuracy over all thresholds.

\subsubsection{Evaluation on ScanNet}

We provide accuracy-threshold plots for the evaluation on the downstream task of relative pose estimation on ScanNet~\cite{dai2017scannet} from which we calculate mAA~\cite{yi2018learning, jin2021image}. We sample a validation set that consists of 16k unique images that form 8k pairs from validation sequences of ScanNet using a sampling procedure similar to~\cite{ono2018lf, wang2020learning}. The hyper-parameter selection and fine-tuning are done in the same way as described in Sec.~\ref{sec:imc_pt}. We estimate the essential matrix using a robust estimator from OpenGV~\cite{kneip2014opengv} with 5000 iterations and a 0.99999 confidence level.

\begin{figure*}
\begin{center}
\begin{subfigure}[b]{0.98\linewidth}
\begin{center}
\includegraphics[width=\linewidth]{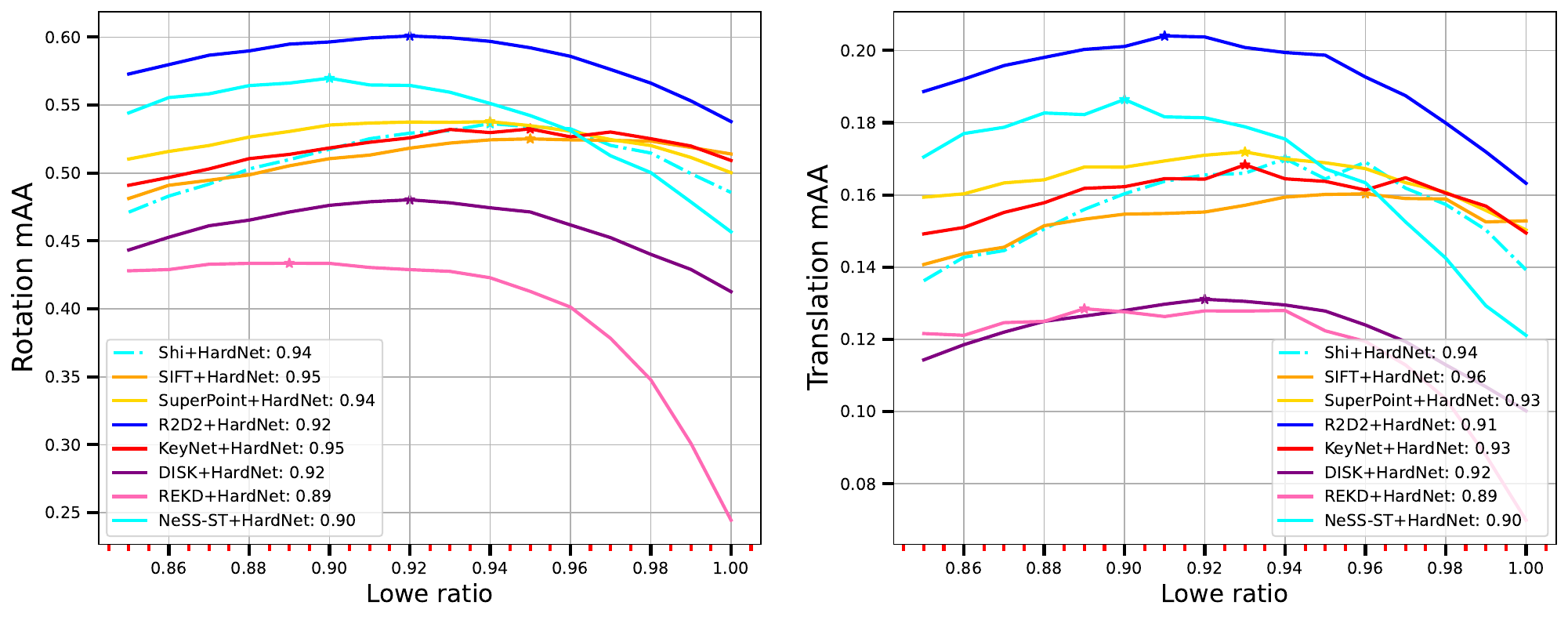}
  \caption{Lowe ratio tuning.}
\end{center}
\end{subfigure}
\begin{subfigure}[b]{0.98\linewidth}
\begin{center}
\includegraphics[width=\linewidth]{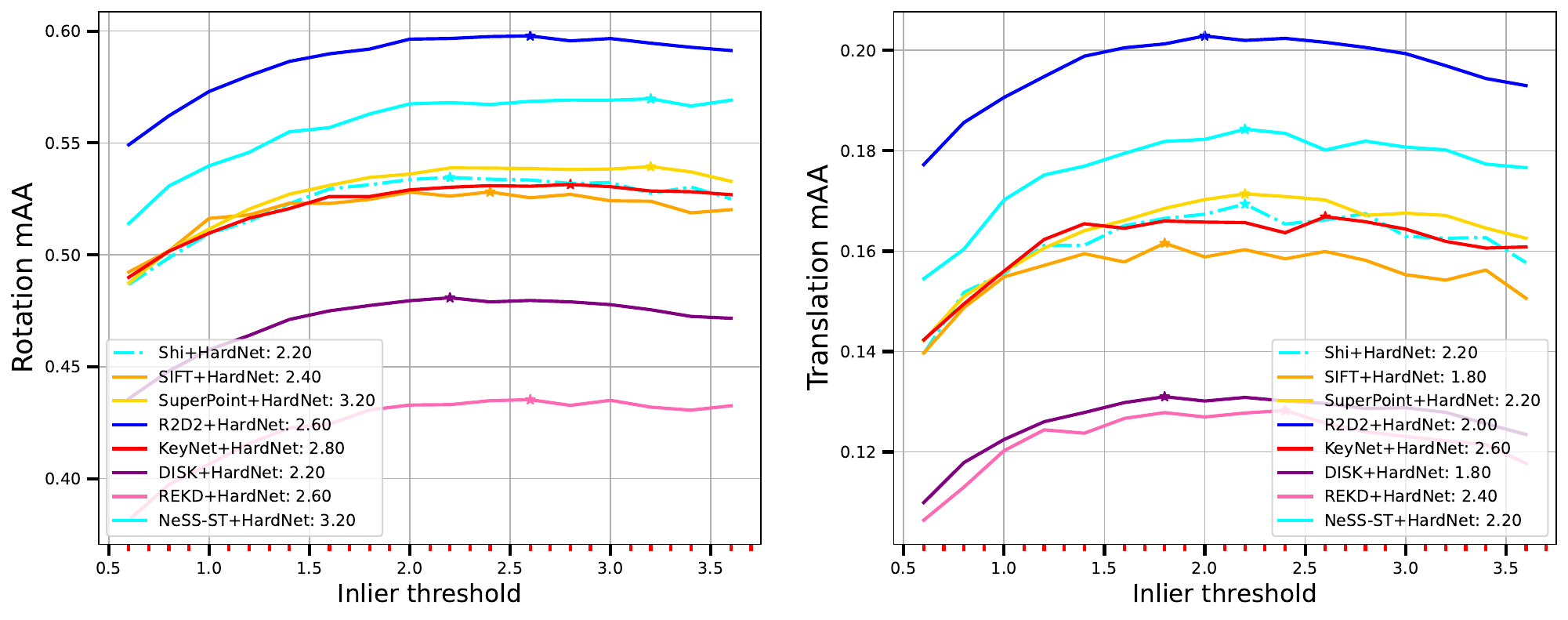}
\caption{Inlier threshold tuning.}
\end{center}
\end{subfigure}
\caption{Hyper-parameter tuning on the validation set of ScanNet~\cite{dai2017scannet} with with 2048 keypoints and full resolution images. We report mAA~\cite{yi2018learning, jin2021image} up to a 10 degrees threshold for rotation and translation for different hyper-parameter values.}
\label{fig:scannet_tuning}
\end{center}
\end{figure*}

Fig.~\ref{fig:scannet} shows that NeSS-ST achieves second place over all thresholds losing only to R2D2~\cite{revaud2019r2d2}. Fig.~\ref{fig:scannet_tuning} presents the hyper-parameters and mAA curves. See Table~\ref{tab:scannet_tuning} for the hyper-parameters used in the evaluation.

\begin{table}
\begin{center}
\resizebox{\columnwidth}{!}{
\begin{tabular}{lcccccc}
\toprule
\multirow{2}{*}{Methods} & Lowe ratio & Inlier threshold\\
\\
\midrule
Shi-Tomasi~\cite{harris1988combined, shi1994good} + HardNet~\cite{mishchuk2017working} & 0.94 & 2.2 \\
SIFT~\cite{lowe2004distinctive} + HardNet~\cite{mishchuk2017working} & 0.95 & 2.0 \\
SuperPoint~\cite{detone2018superpoint} + HardNet~\cite{mishchuk2017working} & 0.93 & 2.2 \\
R2D2~\cite{revaud2019r2d2} + HardNet~\cite{mishchuk2017working} & 0.92 & 2.4 \\
Key.Net~\cite{barroso2019key} + HardNet~\cite{mishchuk2017working} & 0.93 & 2.2(2.6) \\
DISK~\cite{tyszkiewicz2020disk} + HardNet~\cite{mishchuk2017working} & 0.92 & 2.2 \\
REKD~\cite{lee2022self} + HardNet~\cite{mishchuk2017working} & 0.89 & 2.2(2.4) \\
\midrule
NeSS-ST + HardNet~\cite{mishchuk2017working} & 0.9 & 2.2 \\
\bottomrule
\end{tabular}
}
\end{center}
\caption{Hyper-parameters used for essential matrix estimation on ScanNet~\cite{dai2017scannet}. Tuned hyper-parameters, if different, are provided in brackets.}
\label{tab:scannet_tuning}
\end{table}

\subsection{Additional ablation study}

\subsubsection{Base detector ablation}
\label{sec:base_detector_ablation}

We provide accuracy-threshold plots for the ablation on the downstream task of relative pose estimation on IMC-PT~\cite{jin2021image} from which we calculate mAA~\cite{yi2018learning, jin2021image}. Additionally, we report MMA~\cite{mikolajczyk2005performance, dusmanu2019d2} and repeatability~\cite{mikolajczyk2005comparison} for different pixel thresholds on HPatches.

\begin{figure*}
\begin{center}
\includegraphics[width=0.98\linewidth]{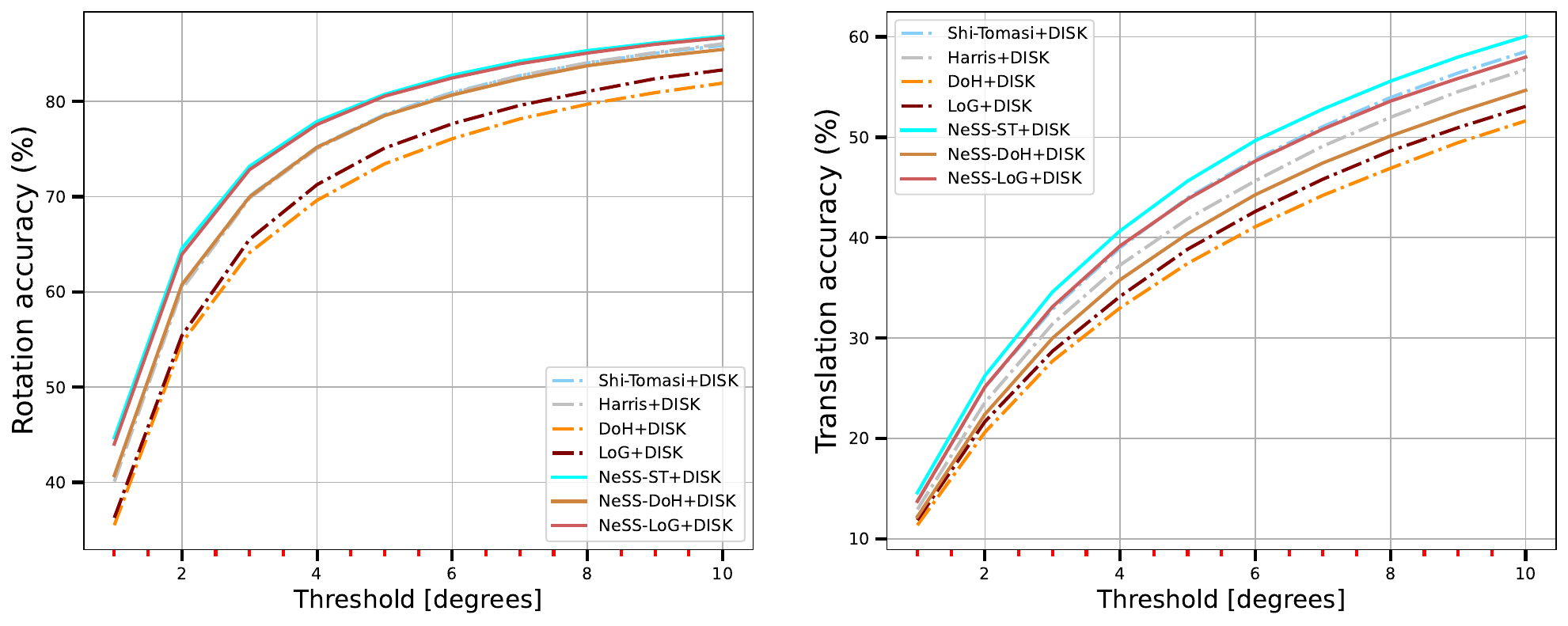}
\end{center}
\caption{Base detector ablation on IMC-PT~\cite{jin2021image} with 2048 keypoints and full resolution images.  We report pose estimation accuracy~\cite{yi2018learning, wang2020learning, jin2021image} in \%.}
\label{fig:imcpt_base_detector}
\end{figure*}

\begin{figure*}
\begin{center}
\begin{subfigure}[b]{0.98\linewidth}
\begin{center}
\includegraphics[width=\linewidth]{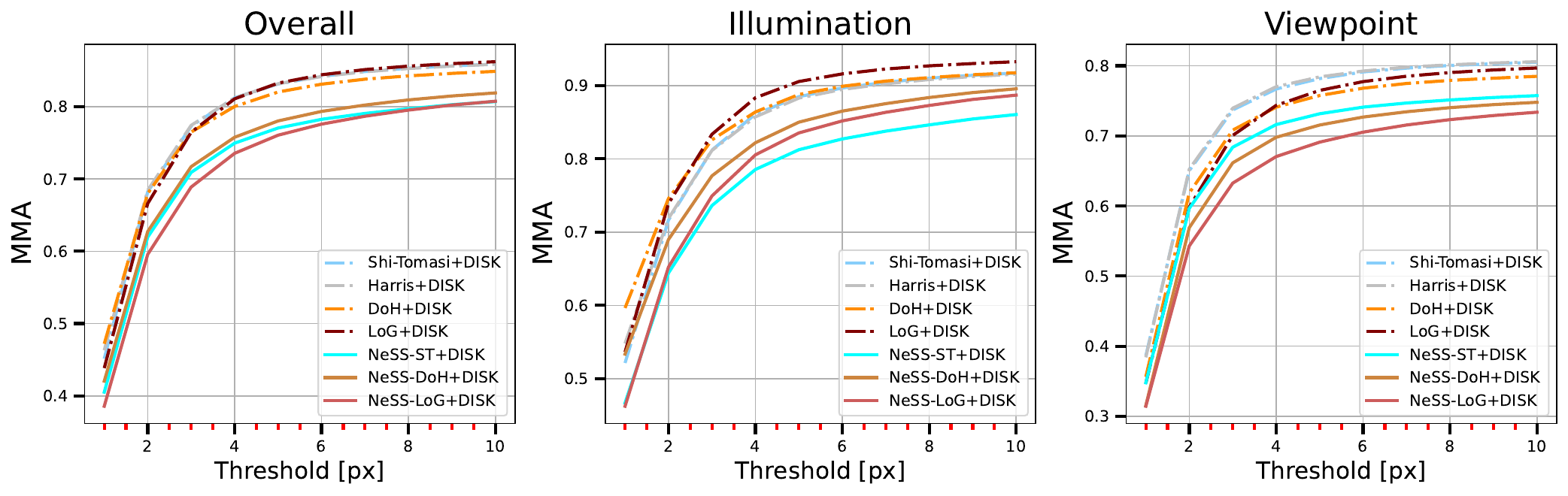}
  \caption{MMA~\cite{mikolajczyk2005performance, dusmanu2019d2}.}
\end{center}
\end{subfigure}
\begin{subfigure}[b]{0.98\linewidth}
\begin{center}
\includegraphics[width=\linewidth]{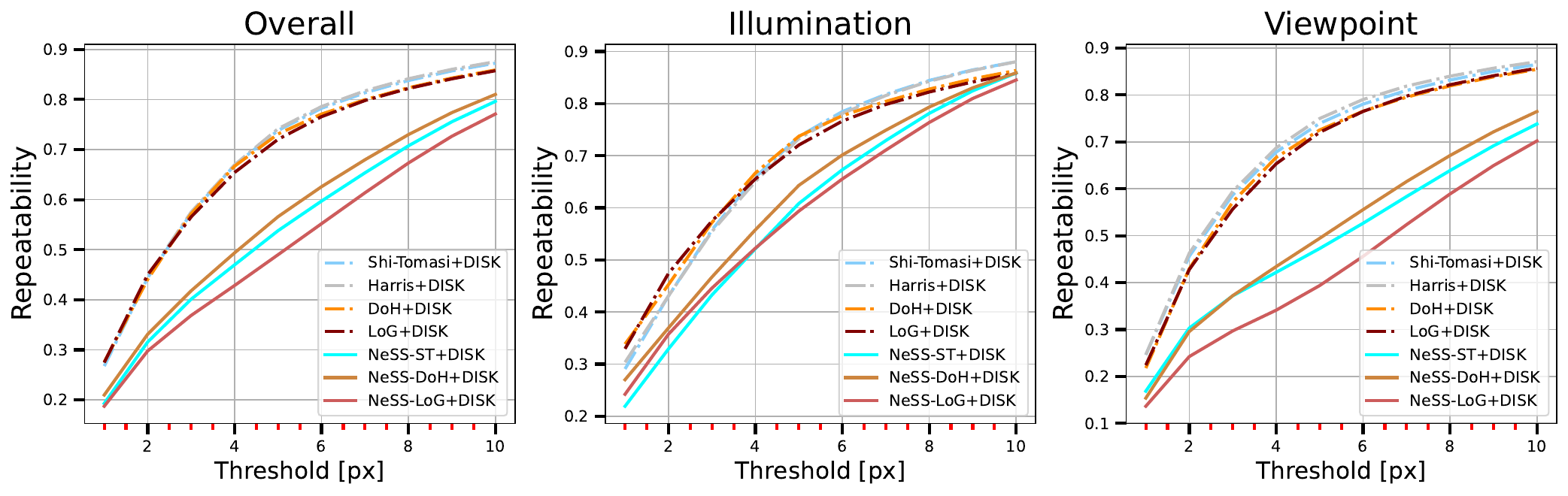}
\caption{Repeatability~\cite{mikolajczyk2005comparison}.}
\end{center}
\end{subfigure}
\caption{Base detector ablation on HPatches~\cite{balntas2017hpatches} with 2048 keypoints and full resolution images. We report classical metrics.}
\label{fig:hpatches_base_detector}
\end{center}
\end{figure*}

Shi-Tomasi~\cite{harris1988combined, shi1994good} shows the best performance on the downstream task among all handcrafted detectors as illustrated in Fig.~\ref{fig:imcpt_base_detector}. In contrast with the downstream task evaluation, Fig.~\ref{fig:hpatches_base_detector} shows that all handcrafted detectors have approximately the same performance in both MMA and repeatability. Using the detectors together with NeSS decreases their classical metrics scores, but, in return, it boots their performance on the downstream task. These results indicate that the conventional metrics cannot comprehensively assess the ability of a detector to perform well in applications.

\subsubsection{Stability score design ablation}

Like in Sec.~\ref{sec:base_detector_ablation}, we provide mAA~\cite{yi2018learning, jin2021image} on IMC-PT~\cite{jin2021image} and classical metrics~\cite{mikolajczyk2005performance, dusmanu2019d2} on HPatches~\cite{balntas2017hpatches}. Additionally, we conduct experiments to showcase the influence of thresholding on SS-ST and RS-ST detectors.

\begin{figure*}
\begin{center}
\includegraphics[width=0.98\linewidth]{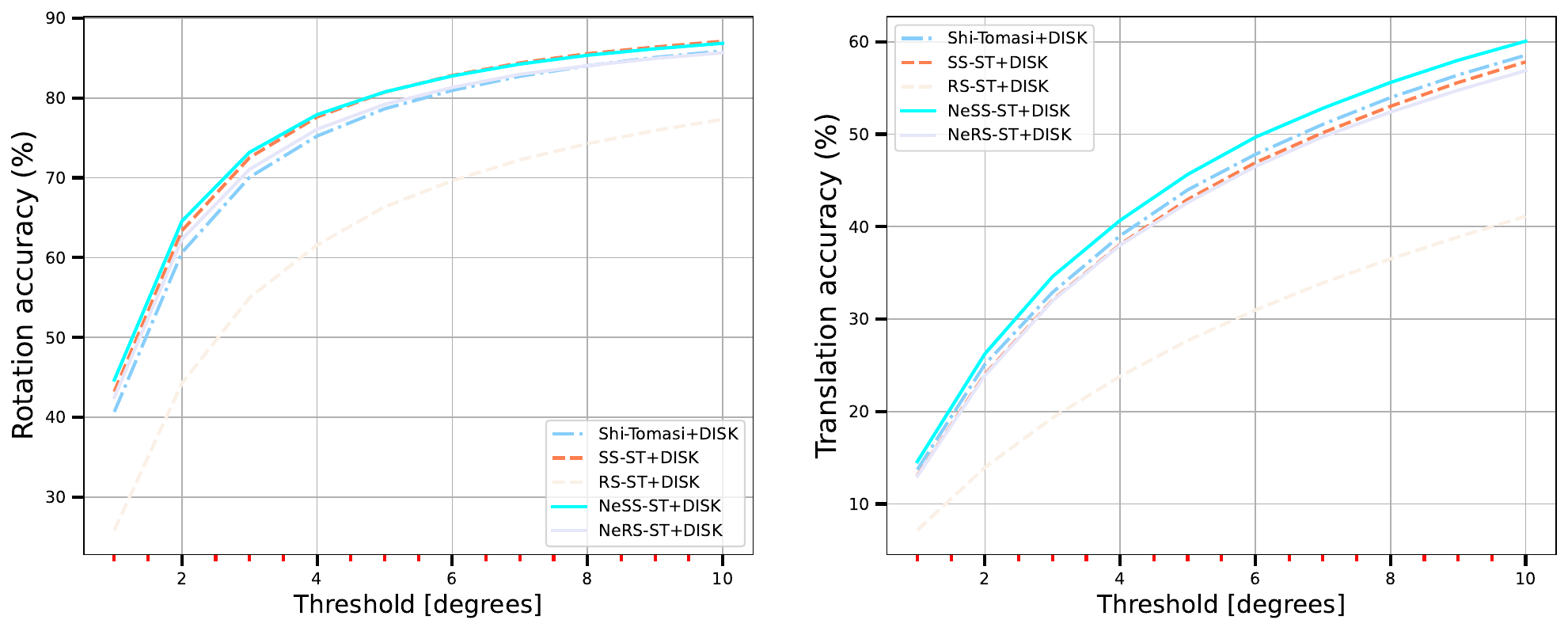}
\end{center}
\caption{SS design ablation on IMC-PT~\cite{jin2021image} with 2048 keypoints and full resolution images.  We report pose estimation accuracy~\cite{yi2018learning, wang2020learning, jin2021image} in \%.}
\label{fig:imcpt_ss_abl}
\end{figure*}

\begin{figure*}
\begin{center}
\begin{subfigure}[b]{0.98\linewidth}
\begin{center}
\includegraphics[width=\linewidth]{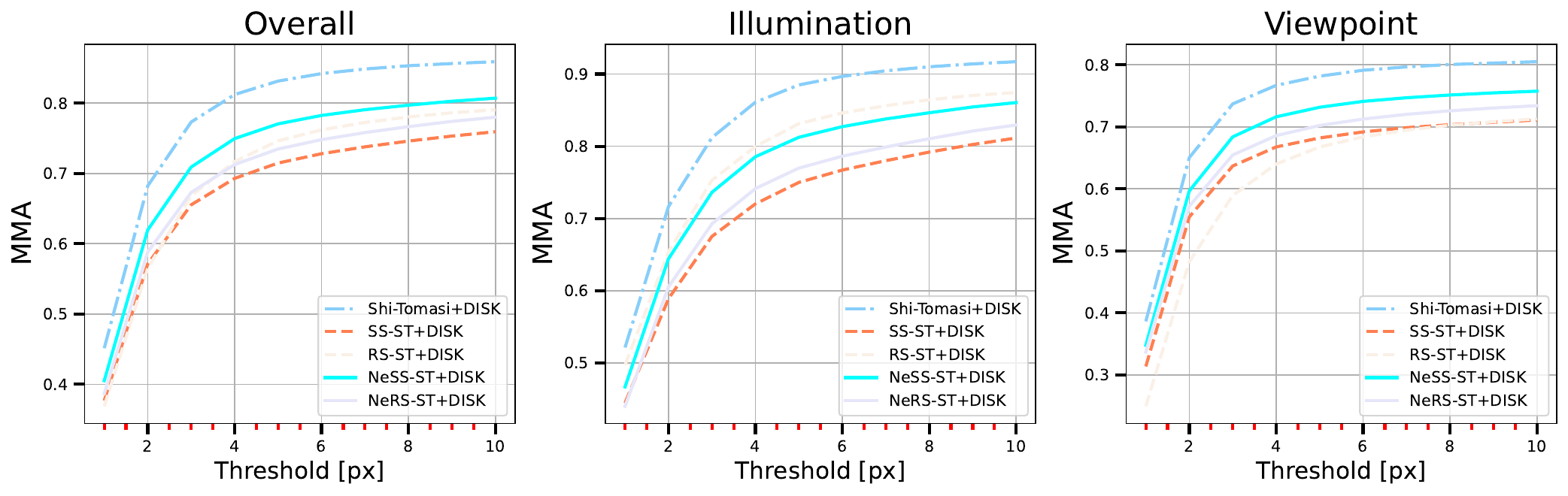}
  \caption{MMA~\cite{mikolajczyk2005performance, dusmanu2019d2}.}
\end{center}
\end{subfigure}
\begin{subfigure}[b]{0.98\linewidth}
\begin{center}
\includegraphics[width=\linewidth]{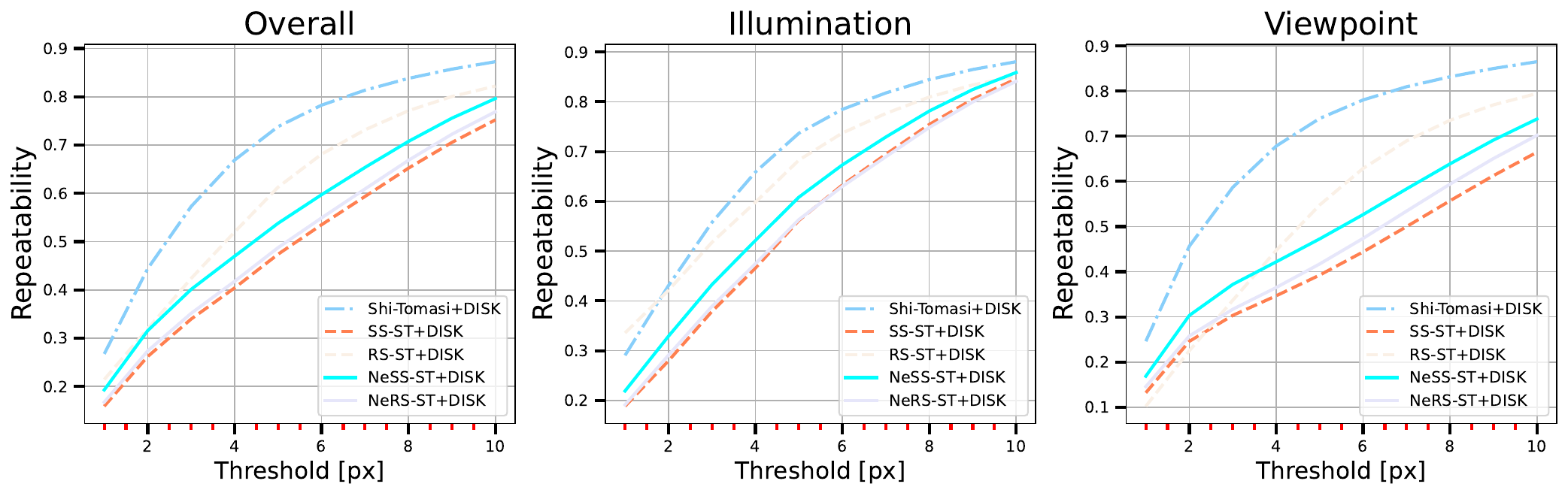}
\caption{Repeatability~\cite{mikolajczyk2005comparison}.}
\label{fig:hpatches_ss_abl_rep}
\end{center}
\end{subfigure}
\caption{SS design ablation on HPatches~\cite{balntas2017hpatches} with 2048 keypoints and full resolution images. We report classical metrics.}
\label{fig:hpatches_ss_abl}
\end{center}
\end{figure*}

\begin{figure*}
\begin{center}
\begin{subfigure}[b]{0.98\linewidth}
\begin{center}
\includegraphics[width=\linewidth]{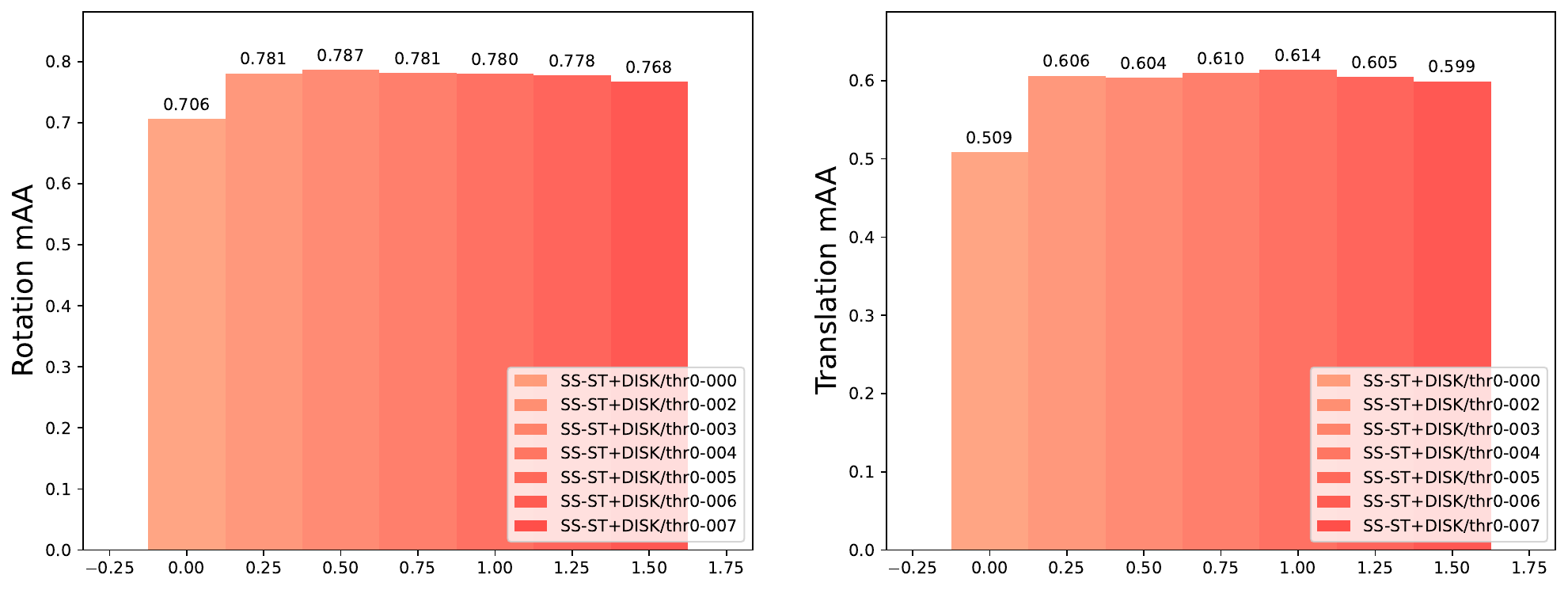}
  \caption{SS-ST evaluated with different values of $t_\mathit{Shi}$ (0.0, 0.002, 0.003, 0.004, 0.005, 0.006 and 0.007).}
\end{center}
\end{subfigure}
\begin{subfigure}[b]{0.98\linewidth}
\begin{center}
\includegraphics[width=\linewidth]{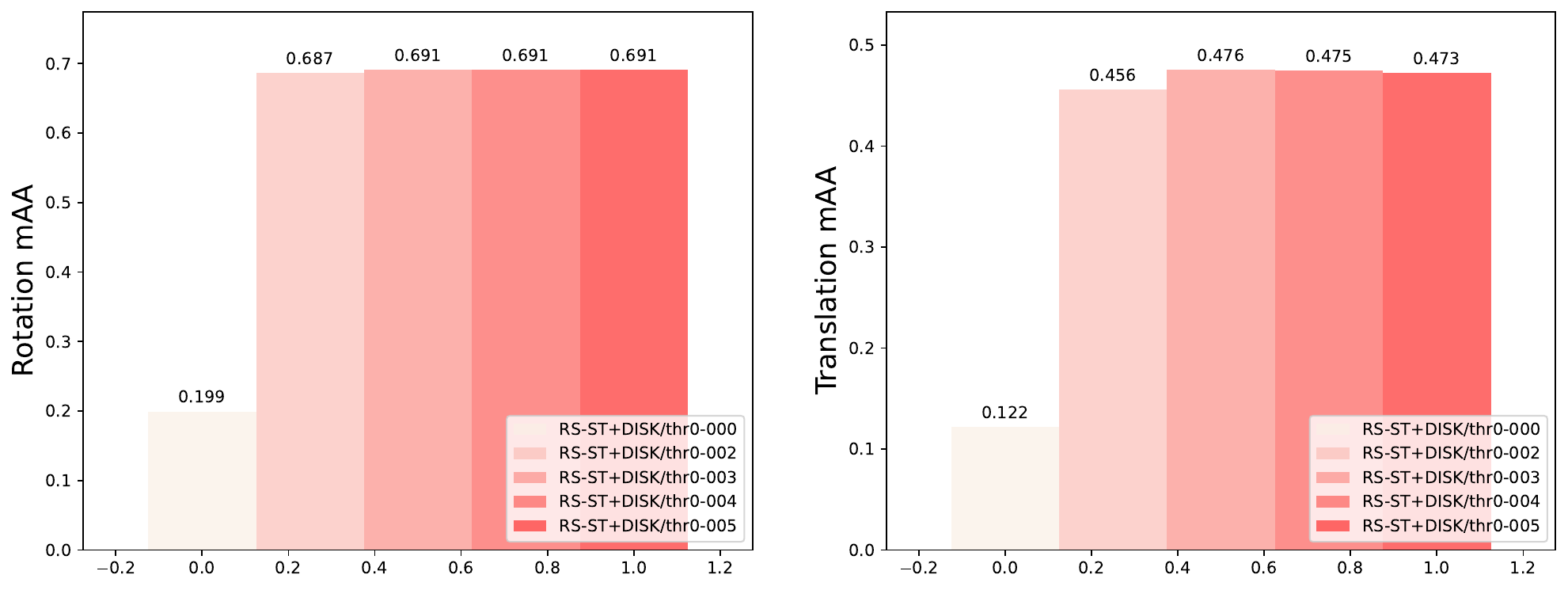}
\caption{RS-ST evaluated with different values of $t_\mathit{Shi}$ (0.0, 0.002, 0.003, 0.004 and 0.005).}
\end{center}
\end{subfigure}
\caption{Ablation of the influence of filtering on performance on the validation set of IMC-PT~\cite{jin2021image} with 2048 keypoints and full resolution images. We report mAA~\cite{yi2018learning, jin2021image} up to a 10 degrees threshold for rotation and translation. }
\label{fig:imcpt_ss_rs_filtering}
\end{center}
\end{figure*}

Fig.~\ref{fig:imcpt_ss_abl} shows that the neural network improves the performance of both SS-ST and RS-ST over all thresholds with models based on the stability score providing better downstream performance. Fig.~\ref{fig:hpatches_ss_abl_rep} illustrates that RS-ST has higher repeatability than SS-ST. Still, the latter performs better on the downstream task. Fig.~\ref{fig:imcpt_ss_rs_filtering} illustrates that thresholding of low-saliency responses is essential for the good performance of both SS-ST and RS-ST.

\end{document}